%% file: evaluation.tex
\documentclass[a4paper,11pt]{article}

\usepackage[a4paper, top=1in, bottom=1.25in, left=1.25in, right=1.25in]{geometry}

\usepackage{placeins}

\usepackage{amsmath}
\usepackage{amsfonts}
\usepackage{xspace}
\usepackage{booktabs}
\usepackage{floatrow}
\def\mytitle{An Evaluation of Change Point Detection Algorithms}

\usepackage{natbib}
\usepackage[hidelinks]{hyperref}
\hypersetup{
    unicode=false,          
    pdftoolbar=true,        
    pdfmenubar=true,        
    pdffitwindow=false,     
    pdfstartview={FitH},    
    pdftitle={\mytitle},    
    pdfauthor={G.J.J. van den Burg and C.K.I. Williams}, 
    pdfcreator={G.J.J. van den Burg},   
    pdfproducer={G.J.J. van den Burg}, 
    pdfkeywords={Change Point Detection, Time Series Analysis, Benchmark Evaluation}, 
    pdfnewwindow=true,      
    colorlinks=true,        
    linkcolor=black,        
    citecolor=black,        
    filecolor=magenta,      
    urlcolor=black           
}

\usepackage[pdftex, dvipsnames]{xcolor}
\usepackage[caption=false]{subfig}
\usepackage{graphicx}
\graphicspath{{./images/}}

\newcommand{\bft}[1]{\mathbf{#1}}

\newcommand{\abs}[1]{\left|#1\right|}
\newcommand{\set}[1]{\left\{#1\right\}}
\newcommand{\given}{\;\vert\;}

\bibpunct{(}{)}{;}{a}{,}{,}

\newif\ifDatasetOverviewAppendix
\makeatletter
\long\def\@makecaption#1#2{
	\vskip 10pt
	\setbox\@tempboxa\hbox{#1: #2}
	\ifdim \wd\@tempboxa >\hsize               
	\ifDatasetOverviewAppendix
	\footnotesize
		#1: #2%
	\else%
		\begin{list}{#1:}{
				\settowidth{\labelwidth}{#1:}
				\setlength{\leftmargin}{\labelwidth}
				\addtolength{\leftmargin}{\labelsep}
			}%
		\item #2%
		\end{list}%
		\par   
	\fi
	\else                                    
	\hbox to\hsize{\hfil\box\@tempboxa\hfil}
	\fi%
}
\makeatother

\newcommand{\MyFigure}[4]{%
	\DatasetOverviewAppendixtrue%
	\setlength{\intextsep}{10pt}
	\begin{figure}[!htbp]
		\centering
		\floatbox[%
		{%
			\capbeside\thisfloatsetup{capbesideposition={left,top},%
				capbesidewidth=5cm}%
		}%
		]{figure}[9cm]%
		{\caption{#4}\label{fig:#2}}%
		{%
			\texttt{\scriptsize{#3}}\par\medskip%
			\includegraphics[width=.8\linewidth]{#1}%
		}
	\end{figure}%
	\DatasetOverviewAppendixfalse%
}

\def\NumRealDatasets{37\xspace}
\def\NumQCDatasets{5\xspace}
\def\NumTotalDatasets{42\xspace}
\def\NumAnnotators{eight\xspace}
\def\NumMethods{14\xspace}
\def\AnnotationsPerDataset{five\xspace}

\begin{document}

\title{\mytitle}
\author{%
	Gerrit J.~J.~van den Burg\thanks{Work done while at The Alan Turing 
		Institute.} \\
	\href{mailto:gertjanvandenburg@gmail.com}{\tt \normalsize 
		gertjanvandenburg@gmail.com}
	\and
	Christopher K.~I.~Williams$^{1,2}$ \\
	\href{mailto:ckiw@inf.ed.ac.uk}{\tt \normalsize ckiw@inf.ed.ac.uk}
}
\date{%
	$^1$ The University of Edinburgh, Edinburgh, UK\\
	$^2$ The Alan Turing Institute, London, UK\\
	\vskip\baselineskip
	\today
}

\maketitle

\begin{abstract}%
	Change point detection is an important part of time series analysis, 
	as the presence of a change point indicates an abrupt and significant 
	change in the data generating process. While many algorithms for 
	change point detection have been proposed, comparatively little  
	attention has been paid to evaluating their performance on real-world 
	time series. Algorithms are typically evaluated on simulated data and 
	a small number of commonly-used series with unreliable ground truth.  
	Clearly this does not provide sufficient insight into the comparative 
	performance of these algorithms. Therefore, instead of developing yet 
	another change point detection method, we consider it vastly more 
	important to properly evaluate existing algorithms on real-world data.  
	To achieve this, we present a data set specifically designed for the 
	evaluation of change point detection algorithms that consists of 
	\NumRealDatasets time series from various application domains.
	Each series was annotated by five human annotators to provide ground 
	truth on the presence and location of change points. We analyze the 
	consistency of the human annotators, and describe evaluation metrics 
	that can be used to measure algorithm performance in the presence of 
	multiple ground truth annotations. Next, we present a benchmark study 
	where \NumMethods algorithms are evaluated on each of the time series 
	in the data set. 
	Our aim is that this data set will serve as a proving ground in the 
	development of novel change point detection algorithms.\\
	\vskip.2\baselineskip
\end{abstract}

\section{Introduction}
\label{sec:intro}

Moments of abrupt change in the behavior of a time series are often cause for 
alarm as they may signal a significant alteration to the data generating 
process. Detecting such \emph{change points} is therefore of high importance 
for data analysts and engineers. The literature on change point detection 
(CPD) is concerned with accurately detecting these events in time series data, 
as well as with modeling time series in which such events occur.  Applications 
of CPD are found in finance \citep{lavielle2007adaptive}, business 
\citep{taylor2018forecasting}, quality control \citep{page1954continuous}, and 
network traffic analysis \citep{kurt2018bayesian}, among many other domains.

There is a large body of work on both the theoretical and practical aspects of 
change point detection. However, less attention has been paid to the 
evaluation of CPD algorithms on real-world time series. Existing work often 
follows a predictable strategy when presenting a novel detection algorithm: 
the method is first evaluated on a set of \emph{simulated} time series with 
known change points where both the model fit and detection accuracy are 
evaluated.  An obvious downside of such experiments is that the dynamics of 
the simulated data are often particular to the paper, and any model that 
corresponds to these dynamics has an unfair advantage. Subsequently the 
proposed method is typically evaluated on only a small number of real-world 
time series.  These series are often reused (e.g.,~the univariate well-log 
data of \citealp{oruanaidh1996numerical}), may have been preprocessed 
(e.g.,~by removing outliers or seasonality), and may not have unambiguous 
ground truth available.
When evaluating a proposed method on real-world data, post-hoc analysis is 
frequently applied to argue why the locations detected by the algorithm 
correspond to known events. Clearly, this is not a fair or accurate approach 
to evaluating novel change point detection algorithms.

Rather than proposing yet another detection method and evaluating it in the 
manner just described, we argue that it is vastly more important to create a 
realistic benchmark data set for CPD algorithms. Such a data set will allow 
researchers to evaluate their novel algorithms in a systematic and realistic 
way against existing alternatives. Furthermore, since change point detection 
is generally an unsupervised learning problem, an evaluation on real-world 
data is the most reliable approach to properly compare detection accuracy. A 
benchmark data set can also uncover important failure modes of existing 
methods that may not occur in simulated data, and suggest avenues for future 
research based on a quantitative analysis of algorithm performance.

Change point detection can be viewed as partitioning an input time series into 
a number of segments. It is thus highly similar to the problem of image 
segmentation, where an image is partitioned into distinct segments based on 
different textures or objects. Historically, image segmentation suffered from 
the same problem as change point detection does today, due to the absence of a 
large, high-quality data set with objective ground truth (as compared to, for 
instance, image classification). It was not until the Berkeley Segmentation 
Data Set \citep[BSDS;][]{martin2001database} that quantitative comparison of 
segmentation algorithms became feasible. We believe that it is essential to 
develop a similar data set for change point detection, to enable both the 
research community as well as data analysts to quantitatively compare CPD 
algorithms.

To achieve this goal, we present the first data set specifically designed to 
evaluate change point detection algorithms that consists of time series from 
diverse application domains. We developed an annotation tool and collected 
annotations from data scientists for \NumRealDatasets real-world time series.  
Using this data set, we perform a large-scale quantitative evaluation of 
existing algorithms. We analyze the results of this comparison to uncover 
failure cases of present methods and discuss various potential improvements.  
To summarize, our main contributions are as follows.
\begin{enumerate}
	\item We describe the design and collection of a data set for change 
		point detection algorithms and analyze the data set for 
		annotation quality and consistency. The data set is made 
		freely available to accelerate research on change point 
		detection.\footnote{See 
			\url{https://github.com/alan-turing-institute/TCPD}.}
	\item We present metrics for the evaluation of change point detection 
		algorithms that take multiple annotations per data set into 
		account.
	\item We evaluate a significant number of existing change point 
		detection algorithms on this data set and provide the first 
		benchmark study on a large number of real-world data sets 
		using two distinct experimental setups.\footnote{All code 
			needed to reproduce our results is made available 
			through an online repository:\\
			\url{https://github.com/alan-turing-institute/TCPDBench}.}
\end{enumerate}

The remainder of this paper is structured as follows. In 
Section~\ref{sec:related_work} we give a high-level overview and 
categorization of existing work on CPD. Section~\ref{sec:evaluation_metrics} 
presents a framework for evaluating change point methods when ground truth 
from multiple human annotators is available.  The annotation tool and data set 
collection process are described in Section~\ref{sec:change_point_dataset}, 
along with an analysis of data quality and annotator consistency. The design 
of the benchmark study and an analysis of the results is presented in 
Section~\ref{sec:benchmark_study}. Section~\ref{sec:discussion} concludes.

\section{Related Work}
\label{sec:related_work}

Methods for CPD can be roughly categorized as follows: (1) online vs.~offline, 
(2) univariate vs.~multivariate, and (3) model-based vs.~nonparametric. We can 
further differentiate between Bayesian and frequentist approaches among the 
model-based methods, and between divergence estimation and heuristics in the 
nonparametric methods. Due to the large amount of work on CPD we only provide 
a high-level overview of some of the most relevant methods here, which also 
introduces many of the methods that we subsequently consider in our evaluation 
study. For a more expansive review of change point detection algorithms, see 
e.g.,~\citet{aminikhanghahi2017survey} or \citet{truong2020selective}. Note 
that we focus here on the unsupervised change point detection problem, without 
access to external data that can be used to tune hyperparameters 
\citep{hocking2013learningsparse,truong2017penalty}.

Let $\bft{y}_t \in \mathcal{Y}$ denote the observations for time steps $t = 1, 
2, \ldots$, where the domain $\mathcal{Y}$ is $d$-dimensional and typically 
assumed to be a subset of $\mathbb{R}^d$. A segment of the series from $t = a, 
a+1, \ldots, b$ will be written as $\bft{y}_{a:b}$. The ordered set of change 
points is denoted by $\mathcal{T} = \{\tau_0, \tau_1, \ldots, \tau_n \}$, with 
$\tau_0 = 1$ for notational convenience. In \emph{offline} change point 
detection we further use $T$ to denote the length of the series and define 
$\tau_{n+1} = T + 1$. Note that we assume that a change point marks the first 
observation of a new segment.

Early work on CPD originates from the quality control literature.  
\citet{page1954continuous} introduced the CUSUM method that detects where the 
corrected cumulative sum of observations exceeds a threshold value.  
Theoretical analysis of this method was subsequently provided by 
\citet{lorden1971procedures}. \citet{hinkley1970inference} generalized this 
approach to testing for differences in the maximum likelihood, i.e.,~testing 
whether $\log p(\bft{y}_{1:T} \given \theta)$ and $\max_{\tau} [ \log 
p(\bft{y}_{1:\tau-1} \given \theta_1) + \log p(\bft{y}_{\tau:T} \given 
\theta_2)]$ differ significantly, for model parameters $\theta$, $\theta_1$, 
and $\theta_2$.  Alternatives from the literature on structural breaks include 
the use of a Chow test \citep{chow1960tests} or multiple linear regression 
\citep{bai2003computation}. 

In offline \emph{multiple} change point detection the optimization problem is 
given by
\begin{equation}
	\label{eq:offline_cpd}
	\min_{\mathcal{T}} \sum_{i=1}^{n+1} 
	\ell(\bft{y}_{\tau_{i-1}:\tau_{i}-1}) + \lambda P(n),
\end{equation}
with $\ell(\cdot)$ a cost function for a segment (e.g.,~the negative maximum 
log-likelihood), $\lambda \geq 0$ a hyperparameter, and $P(n)$ a penalty on 
the number of change points. An early method by \citet{scott1974cluster} 
introduced binary segmentation as an approximate CPD algorithm that greedily 
splits the series into disjoint segments based on the above cost function.  
This greedy method thus obtains a solution in $\mathcal{O}(T\log T)$ time. An 
efficient dynamic programming approach to compute an exact solution for the 
multiple change point problem was presented in \citet{auger1989algorithms}, 
with $\mathcal{O}(m T^2)$ time complexity for a given upper bound $m$ on the 
number of change points.

Improvements to these early algorithms were made by requiring the cost 
function to be additive, i.e.,~$\ell(\bft{y}_{a:b}) = \ell(\bft{y}_{a:\tau-1}) 
+ \ell(\bft{y}_{\tau:b})$. Using this assumption, \citet{jackson2005algorithm} 
present a dynamic programming algorithm with $\mathcal{O}(T^2)$ complexity.  
\citet{killick2012optimal} subsequently extended this method with a pruning 
step on potential change point locations, which reduces the time complexity to 
be approximately linear under certain conditions on the data generating 
process and the distribution of change points. A different pruning approach 
was proposed by \citet{rigaill2015pruned} using further assumptions on the 
cost function, which was extended by \citet{maidstone2017optimal} and made 
robust against outliers by \citet{fearnhead2019changepoint}. Recent work in 
offline CPD also includes the Wild Binary Segmentation method 
\citep{fryzlewicz2014wild} that extends binary segmentation by applying the 
CUSUM statistic to randomly drawn subsets of the time series. Furthermore, the 
Prophet method for Bayesian time series forecasting 
\citep{taylor2018forecasting} supports fitting time series with change points, 
even though it is not a dedicated CPD method. These offline detection methods 
are generally restricted to univariate time series.

Bayesian \emph{online} change point detection was simultaneously proposed by 
\citet{fearnhead2007line} and \citet{adams2007bayesian}. These models learn a 
probability distribution over the ``run length'', which is the time since the 
most recent change point. This gives rise to a recursive message-passing 
algorithm for the joint distribution of the observations and the run lengths.  
The formulation of \citet{adams2007bayesian} has been extended in several 
works to include online hyperparameter optimization \citep{turner2009adaptive} 
and Gaussian Process segment models 
\citep{garnett2009sequential,saatci2010gaussian}. Recent work by 
\citet{knoblauch2018spatio} has added support for model selection and 
spatio-temporal models, as well as robust detection using $\beta$-divergences 
\citep{knoblauch2018doubly}.

Nonparametric methods for CPD are typically based on explicit hypothesis 
testing, where a change point is declared when a test statistic exceeds a 
certain threshold.  This includes kernel change point analysis 
\citep{harchaoui2009kernel}, the use of an empirical divergence measure 
\citep{matteson2014nonparametric}, as well as the use of histograms to 
construct a test statistic \citep{boracchi2018quanttree}. Both the Bayesian 
online change point detection methods and the nonparametric methods support 
multidimensional data.

Finally, we observe that in the field of genomics a number of data sets exist 
that can be used to evaluate change point detection methods 
\citep[e.g.,][]{hocking2013learningsparse,hocking2013learning,hocking2014seganndb}.
While this is an important application domain for CPD algorithms, we are 
interested in evaluating the performance of change point algorithms on 
\emph{general} real-world time series from diverse domains.  Moreover, the 
data sets proposed in \citet{hocking2013learning,hocking2014seganndb} make use 
of labeled \emph{regions} that are annotated as containing a change point or 
not, instead of labels of the exact change point locations. While in this work 
exact labels are preferred over labeled regions, the latter can certainly 
simplify the annotation task for longer time series. We allow a small margin 
of error in the evaluation metrics to take any potential uncertainty of the 
annotators into account.

\section{Evaluation Metrics}
\label{sec:evaluation_metrics}

In this section we review common metrics for evaluating CPD algorithms and 
provide those that can be used to compare algorithms against multiple ground 
truth annotations. These metrics will also be used to quantify the consistency 
of the annotations of the time series, which is why they are presented here.  
Our discussion bears many similarities to that provided by 
\citet{martin2004learning} and \citet{arbelaez2010contour} for the BSDS, due 
to the parallels between change point detection and image segmentation.

Existing metrics for change point detection can be roughly divided between 
clustering metrics and classification metrics. These categories represent 
distinct views of the change point detection problem, and we discuss them in 
turn. The locations of change points provided by annotator $k \in \{1, \ldots, 
K\}$ are denoted by the ordered set $\mathcal{T}_k = \set{\tau_1, \ldots, 
	\tau_{n_k}}$ with $\tau_i \in [1, T]$ for $i = 1, \ldots, n_k$ and 
$\tau_i < \tau_j$ for $i < j$. Then $\mathcal{T}_k$ implies a 
\emph{partition}, $\mathcal{G}_k$, of the interval $[1, T]$ into disjoint sets 
$\mathcal{A}_j$, where $\mathcal{A}_j$ is the segment from $\tau_{j-1}$ to 
$\tau_j - 1$ for $j=1, \ldots, n_k + 1$.  Recall that we use $\tau_0 = 1$ and 
$\tau_{n_k+1} = T + 1$ for notational convenience.

Evaluating change point algorithms using clustering metrics corresponds to the 
view that change point detection inherently aims to divide the time series 
into distinct regions with a constant data generating process. Clustering 
metrics such as the variation of information 
\citep[VI;][]{arabie1973multidimensional}, adjusted Rand index 
\citep[ARI;][]{hubert1985comparing}, Hausdorff distance 
\citep{hausdorff1927mengenlehre}, and segmentation covering metric 
\citep{everingham2010pascal,arbelaez2010contour} can be readily applied. We 
disregard the Hausdorff metric because it is a maximum discrepancy metric and 
can therefore obscure the true performance of methods that report many false 
positives.  Following the discussion in \citet{arbelaez2010contour} regarding 
the difficulty of using VI for multiple ground truth partitions and the small 
dynamic range of the Rand index, we choose to use the covering metric as the 
clustering metric in the experiments. 

For two sets $\mathcal{A}, \mathcal{A}' \subseteq [1, T]$ the Jaccard index, 
also known as Intersection over Union, is given by
\begin{equation}
	J(\mathcal{A}, \mathcal{A}') = \frac{\abs{\mathcal{A} \cap 
			\mathcal{A}'}}{\abs{\mathcal{A} \cup \mathcal{A}'}}.
\end{equation}
Following \citet{arbelaez2010contour}, we define the covering metric of a 
partition $\mathcal{G}$ by a partition $\mathcal{G}'$ as
\begin{equation}
	C(\mathcal{G}, \mathcal{G}') = \frac{1}{T} \sum_{\mathcal{A} \in 
		\mathcal{G}} \abs{\mathcal{A}} \cdot \max_{\mathcal{A}' \in 
		\mathcal{G}'} J(\mathcal{A}, \mathcal{A}').
\end{equation}
For a collection $\set{\mathcal{G}_k}_{k=1}^K$ of ground truth partitions 
provided by the human annotators and a partition $\mathcal{S}$ given by an 
algorithm, we compute the average of $C(\mathcal{G}_k, \mathcal{S})$ for all 
annotators as a single measure of performance.

An alternative view of evaluating CPD algorithms considers change point 
detection as a classification problem between the ``change point'' and 
``non-change point'' classes 
\citep{killick2012optimal,aminikhanghahi2017survey}. Because the number of 
change points is generally a small proportion of the number of observations in 
the series, common classification metrics such as the accuracy score will be 
highly skewed. It is more useful to express the effectiveness of an algorithm 
in terms of \emph{precision} (the ratio of correctly detected change points 
over the number of detected change points) and \emph{recall} (the ratio of 
correctly detected change points over the number of true change points).  The 
$F_{\beta}$-measure provides a single quantity that incorporates both 
precision, $P$, and recall, $R$, through
\begin{equation}
	F_{\beta} = \frac{(1 + \beta^2) P R}{\beta^2 P + R},
\end{equation}
with $\beta = 1$ corresponding to the well-known F1-score 
\citep{van1979information}.

When evaluating CPD algorithms using classification metrics it is common to 
define a margin of error around the true change point location to allow for 
minor discrepancies \citep[e.g.,][]{killick2012optimal,truong2020selective}.  
However with this margin of error we must be careful to avoid double counting, 
so that for multiple detections within the margin around a true change point 
only one is recorded as a true positive \citep{killick2012optimal}. To handle 
the multiple sets of ground truth change points provided by the annotator, we 
follow the precision-recall framework from the BSDS proposed in 
\citet{martin2004learning}. Let $\mathcal{X}$ denote the set of change point 
locations provided by a detection algorithm and let $\mathcal{T}^* = \bigcup_k 
\mathcal{T}_k$ be the combined set of all human annotations. For a set of 
ground truth locations $\mathcal{T}$ we define the set of true positives 
$\text{TP}(\mathcal{T}, \mathcal{X})$ of $\mathcal{X}$ to be those $\tau \in 
\mathcal{T}$ for which $\exists x \in \mathcal{X}$ such that $\abs{\tau - x} 
\leq M$, while ensuring that only one $x \in \mathcal{X}$ can be used for a 
single $\tau \in \mathcal{T}$.
The latter requirement is needed to avoid the double counting mentioned 
previously, and $M \geq 0$ is the allowed margin of error. The precision and 
recall are then defined as
\begin{align}
	P &= \frac{\abs{\text{TP}(\mathcal{T}^*, 
			\mathcal{X})}}{\abs{\mathcal{X}}}, \\
	R &= \frac{1}{K} \sum_{k=1}^K \frac{\abs{\text{TP}(\mathcal{T}_k, 
			\mathcal{X})}}{\abs{\mathcal{T}_k}}.
\end{align}
With this definition of precision, we consider as false positives only those 
detections that correspond to \emph{no} human annotation. Further, by defining 
the recall as the average of the recall for each individual annotator we 
encourage algorithms to explain \emph{all} human annotations without favoring 
any annotator in particular.\footnote{This definition of recall is the 
	\emph{macro}-average of the recall for each annotator, as opposed to 
	the \emph{micro}-average, $R' = (\sum_k \abs{\text{TP}(\mathcal{T}_k, 
		\mathcal{X})})/(\sum_k \abs{\mathcal{T}_k})$. When using $R'$ 
	situations can arise where it is favorable for the detector to agree 
	with the annotator with the \emph{most} annotations, which is not a 
	desirable property.} Note that precision and recall are undefined if 
the annotators or the algorithm declare that a series has no change points.  
To avoid this, we include $t = 1$ as a trivial change point in $\mathcal{T}_k$ 
and $\mathcal{X}$. Since change points are interpreted as the start of a new 
segment this choice does not affect the meaning of the detection results.

Using both the segmentation covering metric and the F1-score in the benchmark 
study allows us to explore both the clustering and classification view of 
change point detection. In the experiments we use a margin of error of $M = 
5$.

\section{Change Point Data Set}
\label{sec:change_point_dataset}

This section describes the tool used in collecting the change point 
annotations, as well as the data sources, the experimental protocol, and an 
analysis of the obtained annotations.

\subsection{Annotation Tool}%
\label{sub:annotation_tool}

\begin{figure}[t]
	\centering
	\fbox{\includegraphics[width=0.5\linewidth]{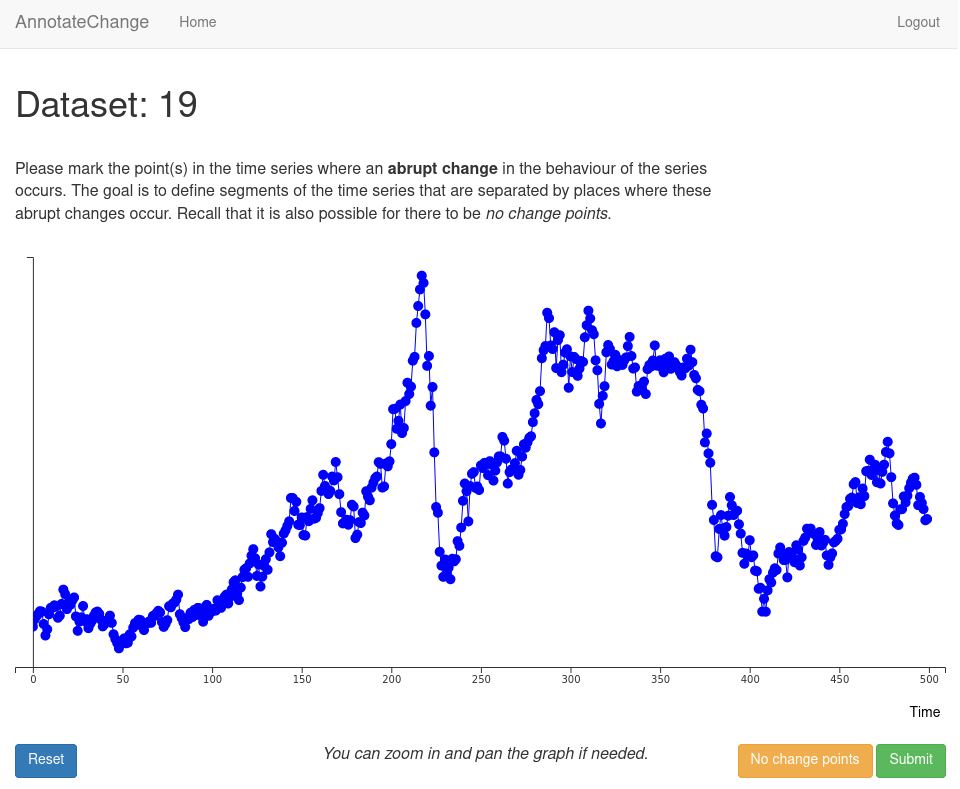}}
	\caption{The annotation tool created to collect time series 
		annotations. Note that no contextual information of the time 
		series is provided to ensure annotators do not use external 
		knowledge during the process. The \emph{rubric} presented to 
		the annotators is also visible.}
	\label{fig:annotatechange}
\end{figure}

To facilitate the collection of the annotated data set, we created a web 
application that enables annotators to mark change points on a plot of a time 
series (see Figure~\ref{fig:annotatechange}).\footnote{The web application is 
	made available as open-source software, see: 
	\url{https://github.com/alan-turing-institute/AnnotateChange}.}  
Annotators can select multiple change points in the data or can signal that 
they believe the time series contains no change points. The application allows 
the annotators to zoom in and pan to particular sections of the time series 
for closer inspection. Before annotators are provided with any real-world time 
series, they are required to go through a set of introductory pages that 
describe the application and show various types of change points (including 
changes in mean, variance, trend, and seasonality).  These introductory pages 
also illustrate the difference between a change point and an outlier.  During 
this introduction the annotator is presented with several synthetic time 
series with known change points and provided with feedback on the accuracy of 
their annotations.  If the annotator's average F1-score is too low on these 
introductory data sets, they are kindly asked to repeat the introduction.  
This ensures that all annotators have a certain baseline level of familiarity 
with annotating change points.

To avoid that annotators use their knowledge of historical events or are 
affected by other biases, no dates are shown on the time axis and no values 
are present on the vertical axis. The names of the series are also not shown 
to the annotators. Furthermore, the annotation page always displays the 
following task description to the annotators:

\begin{quote}
	Please mark the point(s) in the time series where an \textbf{abrupt 
		change} in the behavior of the series occurs. The goal is to 
	define segments of the time series that are separated by places where 
	these abrupt changes occur. Recall that it is also possible for there 
	to be \emph{no change points}.
\end{quote}

This description is purposefully left somewhat vague to avoid biasing the 
annotators in a specific direction, and is inspired by the rubric in the BSDS 
\citep{martin2001database}.

\subsection{Data Sources}%
\label{sub:dataset_sources}

Time series were collected from various online sources including the 
WorldBank, EuroStat, U.S.~Census Bureau, GapMinder, and Wikipedia. A number of 
time series show potential change point behavior around the financial crisis 
of 2007--2008, such as the price for Brent crude oil 
(Figure~\ref{fig:annotatechange}), U.S.~business inventories, and 
U.S.~construction spending, as well as the GDP series for various countries.  
Other time series are centered around introduced legislation, such as the 
mandatory wearing of seat belts in the U.K., the Montreal Protocol regulating 
CFC emissions, or the regulation of automated phone calls in the U.S.  Various 
data sets are also collected from existing work on change point detection and 
structural breaks, such as the well-log data set 
\citep{oruanaidh1996numerical}, a Nile river data set \citep{durbin2012time}, 
and a sequence from the bee-waggle data set \citep{oh2008learning}. The main 
criterion for including a time series was whether it displayed interesting 
behavior that might include an abrupt change.  Several time series were 
included that may not in fact contain a change point but are nevertheless 
considered interesting for evaluating CPD algorithms due to features such as 
seasonality or outliers.

A total of \NumRealDatasets real time series were collected for the change 
point data set. This includes 33 univariate and 4 multivariate series (with 
either 2 or 4 dimensions). One of the univariate time series 
(\verb+uk_coal_employ+, Figure~\ref{fig:ukcoalemploy} in 
Appendix~\ref{app:datasets}) has two missing observations, and nine other 
series show seasonal patterns. Unbeknownst to the annotators we added  
\NumQCDatasets simulated ``quality control'' series with known change points 
in the data set, which allows us to evaluate the quality of the annotators 
\emph{in situ} (see below). Four of these series have a change point with a 
shift in the mean and the fifth has no change points. Of the quality control 
series with a change point two have a change in noise distribution, one 
contains an outlier, and another has multiple periodic components. The change 
point data set thus consists of \NumTotalDatasets time series in total. The 
average length of the time series in the data set is 
	\input{./analysis_output/constants/SeriesLengthMean.tex}\xspace
, with a minimum of %
	\input{./analysis_output/constants/SeriesLengthMin.tex}\xspace
 and 
a maximum of %
	\input{./analysis_output/constants/SeriesLengthMax.tex}\xspace
. Additional descriptive statistics are 
provided in Table~\ref{tab:descriptive} in Appendix~\ref{app:tables} and 
visualizations of all series are provided in Appendix~\ref{app:datasets}. Note 
that because we ask the annotators to mark \emph{all} points in the series 
that they believe to be change points, we have chosen time series for which 
this task is feasible.

\subsection{Annotation Collection}%
\label{sub:annotation_collection}

Annotations were collected in several sessions by inviting data scientists and 
machine learning researchers to annotate the series. We deliberately chose not 
to ask individuals who have extensive background knowledge on the various 
domains of the time series, as this could affect the objectivity of their 
annotations. For example, asking someone who is familiar with oil prices to 
annotate the \verb+brent_spot+ series might allow them to use outside 
knowledge that is not available from the behavior of the series alone. The 
annotation tool randomly assigned time series to annotators whenever they 
requested a new series to annotate, with a bias towards series that already 
had annotations. The same series was never assigned to an annotator more than 
once. In total, \NumAnnotators annotators provided annotations for the 
\NumTotalDatasets time series, with \AnnotationsPerDataset annotators assigned 
to each time series. On average the annotators marked 
	\input{./analysis_output/constants/UniqueAnnotationsMean.tex}\xspace
 unique change points per series, with a 
standard deviation of %
	\input{./analysis_output/constants/UniqueAnnotationsStd.tex}\xspace
, a minimum of 
	\input{./analysis_output/constants/UniqueAnnotationsMin.tex}\xspace
 and a maximum of 
	\input{./analysis_output/constants/UniqueAnnotationsMax.tex}\xspace
. The provided annotations for each series are 
shown in Appendix~\ref{app:datasets}.

Although all annotators successfully completed the introduction before 
annotating any real time series, some nonetheless commented on the difficulty 
of deciding whether a series had a change point or not. In some cases this was 
due to a significant change spanning multiple time steps, which led to 
ambiguity about whether the transition period should be marked as a separate 
segment or not (see e.g.,~Figure~\ref{fig:annotatechange}). Another source of 
ambiguity were periodic time series with abrupt changes (e.g.,~\texttt{bank}, 
Figure~\ref{fig:bank}), which could be regarded as a stationary switching 
distribution without abrupt changes or a series with frequent change points.  
In these cases annotators were advised to use their intuition and experience 
as data scientists to evaluate whether they considered a potential change to 
be an \emph{abrupt} change. Note that uncertainty in the exact position of an 
annotation is partially alleviated by the margin of error included in the 
evaluation metrics (in particular the F1-score, see 
Section~\ref{sec:evaluation_metrics}).

\subsection{Consistency}%
\label{sub:consistency}

\begin{figure}[tb]
	\def\HistogramWidth{0.47\linewidth}
	\centering
	\subfloat[][Covering Metric]{%
		\centering%
		\includegraphics[width=\HistogramWidth]{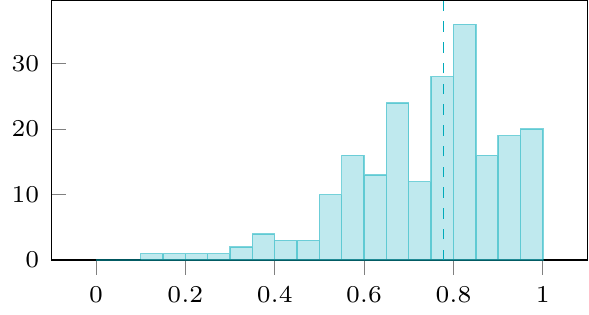}
		\label{fig:anno_hist_cover}
	}
	\quad
	\subfloat[][F1-score]{%
		\centering%
		\includegraphics[width=\HistogramWidth]{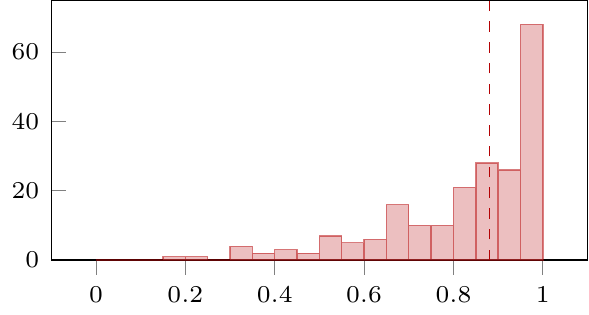}
		\label{fig:anno_hist_f1}
	}
	\caption{Histograms of one vs.~rest annotator performance for both  
		metrics. Median scores over all data sets are shown with dashed 
		lines.}
	\label{fig:anno_hist}
\end{figure}

As mentioned above we included five ``quality control'' series to evaluate the 
accuracy and consistency of the human annotators in practice, which are shown 
in Appendix~\ref{sub:quality_control}. On two of the four series with known 
change points (Figure~\ref{fig:qualitycontrol1} and 
Figure~\ref{fig:qualitycontrol3}) all annotators identified the true change 
point within three time steps of the correct location. On another time series, 
\verb+quality_control_2+, four of the five annotators identified the change 
point within two time steps, with a fifth annotator believing there to be no 
change points (Figure~\ref{fig:qualitycontrol2}). The most difficult series to 
annotate was the one with periodic features (\verb+quality_control_4+, 
Figure~\ref{fig:qualitycontrol4}). Here, two of the annotators voted there to 
be no change points while the three other annotators identified the change 
correctly within three time steps, with one also incorrectly marking several 
other points as change points. The series without change points, 
\verb+quality_control_5+, was correctly identified as such by all annotators.  
This suggests that human annotators can identify the presence and location of 
change points with high accuracy on most kinds of time series, with seasonal 
effects posing the biggest challenge.

To quantitatively assess the consistency of the annotators we evaluate each 
individual annotator against the four others on every time series, using the 
metrics presented previously. Figure~\ref{fig:anno_hist} shows the histograms 
of these one-vs-rest (OvR) scores for both metrics. The figures show that on 
average there is a high degree of agreement between annotators, with the 
median annotator score approximately 0.8 for the covering metric and 0.9 for 
the F1-score (for both metrics a score of 1.0 indicates perfect agreement). It 
can also be seen that the values are generally higher on the F1-score, which 
is due to the margin of error in this metric and the fact that the F1-score 
uses the union of annotated change points in the definition of the precision.  
In Appendix~\ref{app:annotator_agreement} we additionally present a simulation 
study to explore the extent to which the observed annotator agreement can be 
expected by chance. We find that for both metrics and for the majority of time 
series the observed average OvR annotator agreement is significantly higher 
than can be expected by chance. The few cases with low agreement are often due 
to one or two annotators disagreeing with the others. An example of this is 
the \verb+lga_passengers+ series, where there is a high level of agreement 
between four of the annotators about the number and approximate location of 
the change points, but one annotator believes there are no change points (see 
Figure~\ref{fig:lgapassengers}).

\section{Benchmark Study}%
\label{sec:benchmark_study}
This section presents a benchmark study of existing change point algorithms on 
the data set of human-annotated time series. We describe the methods included 
in the study, the setup of the experiments, and the results of the evaluation. 
The code and data used in this benchmark study are available through an online 
repository.\footnote{See: 
	\url{https://github.com/alan-turing-institute/TCPDBench}.}

\begin{table}[t]
	\footnotesize
	\centering
	\caption{The change point detection algorithms considered in the 
		benchmark study.\label{tab:methods}}
	\begin{tabular}{lrr}
		Name & Method & Reference \\
		\hline
		\textsc{amoc} & At Most One Change & 
		\citet{hinkley1970inference} \\
		\textsc{binseg} & Binary Segmentation & 
		\citet{scott1974cluster} \\
		\textsc{bocpd} & Bayesian Online Change Point Detection & 
		\citet{adams2007bayesian} \\
		\textsc{bocpdms} & \textsc{bocpd} with Model Selection & 
		\citet{knoblauch2018spatio} \\
		\textsc{cpnp} & Nonparametric Change Point Detection & 
		\citet{haynes2017computationally} \\
		\textsc{ecp} & Energy Change Point & 
		\citet{matteson2014nonparametric} \\
		\textsc{kcpa} & Kernel Change-Point Analysis & 
		\citet{harchaoui2009kernel} \\
		\textsc{pelt} & Pruned Exact Linear Time & 
		\citet{killick2012optimal} \\
		\textsc{prophet} & Prophet & \citet{taylor2018forecasting} \\
		\textsc{rbocpdms} & Robust \textsc{bocpdms} & 
		\citet{knoblauch2018doubly} \\
		\textsc{rfpop} & Robust Functional Pruning Optimal 
		Partitioning & \citet{fearnhead2019changepoint} \\
		\textsc{segneigh} & Segment Neighborhoods & 
		\citet{auger1989algorithms} \\
		\textsc{wbs} & Wild Binary Segmentation & 
		\citet{fryzlewicz2014wild} \\
		\textsc{zero} & No Change Points & \\
		\hline
	\end{tabular}
\end{table}

\subsection{Experimental Setup}%
\label{sub:methods}

The aim of the study is to obtain an understanding of the performance of 
existing change point methods on real-world data. With that in mind we 
evaluate a large selection of methods that are either frequently used or that 
have been recently developed. Table~\ref{tab:methods} lists the methods 
included in the study. To avoid issues with implementations we use existing 
software packages for all methods, which are almost always made available by 
the authors themselves. We also include the baseline \textsc{zero} method, 
which always returns that a series contains no change points.

Many of the methods have (hyper)parameters that affect the predicted change 
point locations. To obtain a realistic understanding of how accurate the 
methods are in practice, we create two separate evaluations: one showing the 
performance using the \emph{default} settings of the method as defined by the 
package documentation, and one reporting the \emph{maximum} score over a grid 
search of  parameter configurations. We refer to these settings as the 
``Default'' and ``Oracle'' experiments, respectively. The Default experiment 
aims to replicate the common practical setting of a data analyst trying to 
detect potential change points in a time series without any prior knowledge of 
reasonable parameter choices. The Oracle experiment aims to identify the 
highest  possible performance of each algorithm by running a full grid search 
over its hyperparameters. It should be emphasized that while the Oracle 
experiment is important from a theoretical point of view, it is not 
necessarily a realistic measure of practical algorithm performance. Due to the 
unsupervised nature of the change point detection problem we can not in 
general expect those who use these methods to be able to tune the parameters 
such that the results match the \emph{unknown} ground truth. For a detailed 
overview of how each method was applied, see 
Appendix~\ref{app:simulation_details}. 

Since not all methods support multidimensional time series or series with 
missing values, we separate the analysis further into univariate and 
multivariate time series. While it is possible to use the univariate methods 
on multidimensional time series by, for instance, combining the detected 
change points of each dimension, we chose not to do this as it could skew the 
reported performance with results on series that the methods are not 
explicitly designed to handle. Furthermore, because the programming language 
differs between implementations we do not report on the computation time of 
the methods. All series are standardized to zero mean and unit variance to 
avoid issues with numerical precision arising for some methods on some of the 
time series. The quality control time series were not included in the 
benchmark study, but performance on these series is reported in 
Appendix~\ref{app:tables}.

The Bayesian online change point detection algorithms that follow the 
framework of \citet{adams2007bayesian} compute a full probability distribution 
over the location of the change points and therefore do not return a single 
segmentation of the series. To allow for comparisons to the other methods we 
use the maximum a posteriori (MAP) segmentation of the series \citep[see, 
e.g.,][]{fearnhead2007line}. The MAP segmentation results in a single set of 
detected change points that can be compared to those provided by offline 
methods.

\subsection{Results}%
\label{sub:results}

\begin{table}[tb]
	\centering
	\small
	\caption{Scores for both experiments on each metric, separated among 
		univariate and multivariate time series. For the Default 
		experiment the value corresponds to the average of a single 
		run of the algorithm, whereas for the Oracle experiment it is 
		the average of the highest score obtained for each series.  
		Leading methods that do not differ significantly from the best 
		performing one are highlighted in bold. \label{tab:avgscores}}
	\input{./analysis_output/tables/aggregate_scores_wide.tex}
\end{table}

We first analyze the average scores for each of the methods in the two 
experiments, before presenting an in-depth analysis of statistically 
significant differences. Table~\ref{tab:avgscores} shows the average scores on 
each of the considered metrics for both experiments (complete results are 
available in Appendix~\ref{app:additional_tables_figures}).\footnote{For the 
	Default experiment the \textsc{rbocpdms} method failed on two of the 
	univariate time series (\texttt{bitcoin} and 
	\texttt{scanline\_126007}) and therefore received a score of zero on 
	these series. Results on \texttt{uk\_coal\_employ} are excluded from 
	the results for both experiments because the majority of methods do 
	not handle time series with missing values.} We see that for the 
Default experiment the \textsc{binseg} method of \citet{scott1974cluster} 
achieves the highest average performance on the univariate time series, 
closely followed by \textsc{amoc} (on the covering metric) and \textsc{pelt} 
(on the F1 score).  For this experiment the \textsc{bocpdms} method 
\citep{knoblauch2018spatio} does best for the multivariate series when using 
the covering metric, but \textsc{bocpd} \citep{adams2007bayesian} does best on 
the F1-score. In the Oracle experiment --- where hyperparameters of the 
methods are varied to achieve better performance --- we see that 
\textsc{bocpd} does best for both the univariate and multivariate time series 
on the F1 score, but \textsc{rfpop} does slightly better on the univariate 
experiments when using the covering metric.

We note that the \textsc{zero} method outperforms many of the other methods, 
especially in the Default experiment. This is in part due to the relatively 
small number of change points per series, but also suggests that many methods 
return a large number of false positives. Furthermore, we see that for 
\textsc{prophet} and \textsc{wbs} the hyperparameter tuning has a relatively 
small effect on performance. This is unsurprising however, since these methods 
also have a small number of hyperparameters that can be varied. For instance, 
for \textsc{prophet} only the maximum number of change points can be tuned, 
and the results show that this is insufficient to achieve competitive 
performance. By contrast, parameter tuning has a significant effect on the 
performance of \textsc{kcpa} and \textsc{rfpop}, making the latter competitive 
with the best performing methods. We reiterate however, that the Oracle 
experiment is not indicative of real-world performance, since a data scientist 
would have no automatic way of identifying the most accurate segmentation of 
the series among those found by a grid search (which in the case of 
\textsc{rfpop} total over 2400 configurations, see 
Appendix~\ref{app:simulation_details}).

\begin{figure}[tb]
	\def\RankFigWidth{0.47}
	\centering
	\subfloat[][Default -- covering metric]{%
		\centering
		\includegraphics[width=\RankFigWidth\linewidth]{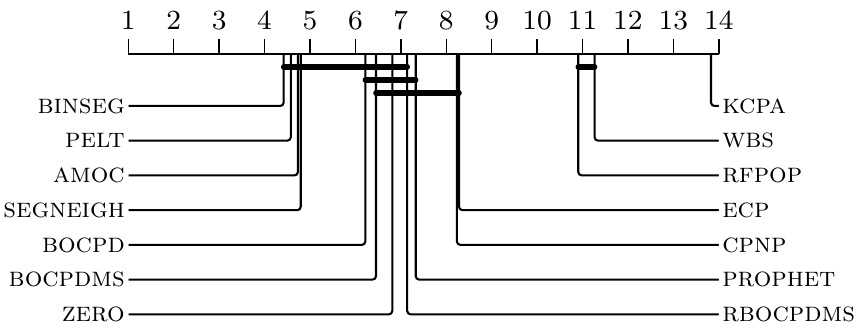}
		\label{fig:cd_default_cover}
	}%
	\quad
	\subfloat[][Default -- F1-score]{%
		\centering
		\includegraphics[width=\RankFigWidth\linewidth]{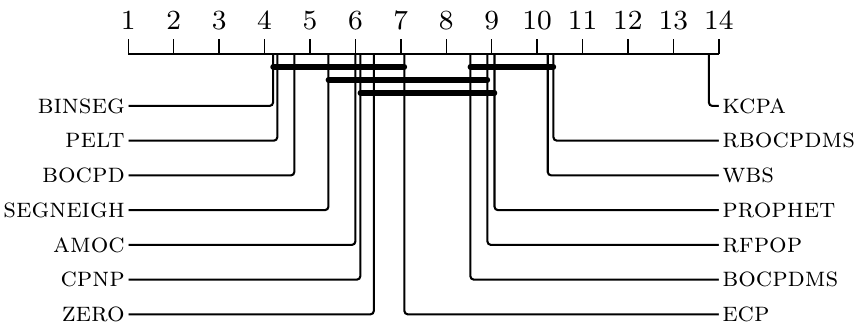}
		\label{fig:cd_default_f1}
	}%
	\\
	\subfloat[][Oracle -- covering metric]{%
		\centering
		\includegraphics[width=\RankFigWidth\linewidth]{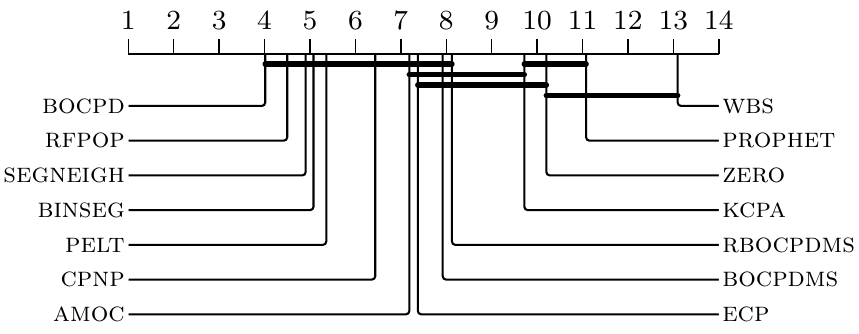}
		\label{fig:cd_oracle_cover}
	}%
	\quad
	\subfloat[][Oracle -- F1-score]{%
		\centering
		\includegraphics[width=\RankFigWidth\linewidth]{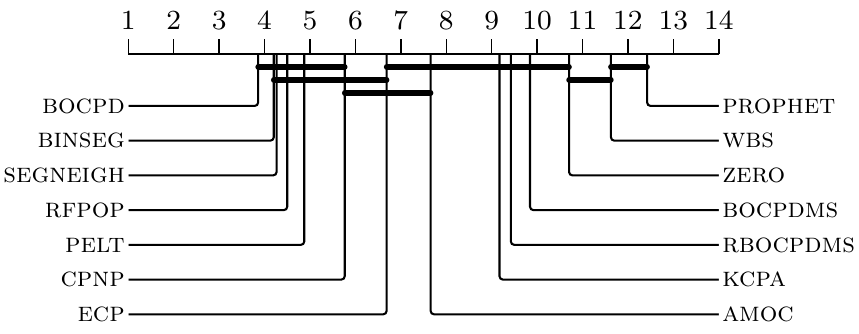}
		\label{fig:cd_oracle_f1}
	}
	\caption{Critical-difference diagrams showing the mean rank of the 
		methods for both experiments and both metrics on univariate 
		time series. Methods that are not significantly different (at 
		$\alpha = 0.05$) are connected. Lower ranks are better.}
	\label{fig:cd_diagrams}
\end{figure}

While the average scores allow for general statements on the performance of 
the different methods, they are not suitable for evaluating the significance 
of performance differences because they represent measurements on distinct 
data sets \citep{demvsar2006statistical}. Instead, an analysis of \emph{ranks} 
can be used to show statistically significant differences. This involves 
assigning a numeric rank $R_{\ell,n}$ to method $\ell$ on data set $n$ with 
the best performing method receiving rank 1, the second best rank 2, etc.  In 
the case of ties fractional ranks are assigned (i.e.,~if two methods achieve 
the highest score on a data set they are both assigned rank 1.5 and the 
third-best method is assigned rank 3). As discussed in detail in 
\citet{demvsar2006statistical}, the Friedman test is a non-parametric test to 
evaluate the null hypothesis that all methods perform equally well.  
Furthermore, pairwise tests can be performed to assess whether two methods 
perform significantly differently, while correcting for the family-wise error 
rate.

Let $\bar{R}_{\ell}$ denote the average rank of method $\ell \in \{1, \ldots, 
L\}$ over all $N$ data sets. Under the null hypothesis of equal performance 
between the methods the Friedman test statistic, given by
\begin{equation}
	\chi_{F}^2 = \frac{12N}{L(L+1)} \left[ \sum_{\ell =1}^L R_{\ell}^2 - 
		\frac{L(L+1)^2}{4} \right],
\end{equation}
follows a $\chi^2$ distribution with $L - 1$ degrees of freedom  
\citep{friedman1937use,friedman1940comparison}. When running this test for the 
experiments on univariate data sets we reject the null hypothesis for both 
metrics on both experiments ($\chi^2 \approx 200$ in all cases, resulting in 
$p$-values of zero). We can subsequently perform post-hoc tests to uncover 
statistically significant differences between methods. While 
\citet{demvsar2006statistical} recommends the Nemenyi test 
\citep{nemenyi1963distribution} as a post-hoc test for individual differences, 
later work by \citet{benavoli2016should} argued that the Wilcoxon signed-rank 
test is to be preferred \citep{wilcoxon1945individual}. Thus, for each pair of 
methods we compute the Wilcoxon signed-rank test with the null hypothesis of 
no significant differences in the performance of two methods. To correct for 
multiple testing we subsequently apply Holm's procedure to control the 
family-wise error rate \citep{holm1979simple}.

To summarize the results of the statistical tests we use the 
critical-difference (CD) diagrams proposed by \citet{demvsar2006statistical}, 
see Figure~\ref{fig:cd_diagrams}. These figures show the average ranks of the 
methods, with methods that are not significantly different from each other 
connected by horizontal lines. For example, in 
Figure~\ref{fig:cd_default_cover} the \textsc{wbs} and \textsc{rfpop} methods 
perform statistically significantly different from all other methods, but not 
from each other. The CD-diagram for the Oracle experiment also shows that when 
hyperparameters can be varied the \textsc{bocpd} method achieves the highest 
average rank. Moreover, depending on the metric considered it also 
statistically significantly outperforms a number of alternative methods (8 
methods for the F1 score).  It is noteworthy that even though \textsc{bocpd} 
assumes a simple Gaussian model with constant mean for each segment, it 
outperforms methods that allow for more complex time series behavior such as 
trend and seasonality (\textsc{prophet}), or autoregressive structures 
(\textsc{bocpdms}). From the critical difference diagrams we can also see that 
for both experiments and metrics, the \textsc{kcpa}, \textsc{prophet}, and 
\textsc{wbs} methods perform statistically significantly worse than the best 
performing methods. We also see that in our experiments no method performs 
significantly better than the \textsc{zero} method on the Default experiment, 
while this does not hold for the Oracle experiment.

By analyzing the performance of the methods on each individual time series in 
the data set we can gain some understanding of which time series are ``easy'', 
in the sense that many methods achieve high scores on the evaluation metrics 
(see Appendix~\ref{app:additional_tables_figures}). For the Default 
experiment, it is particularly noticeable that 9 of the 14 methods achieve an 
F1 score of 1.0 on the \verb+nile+ time series. One possible explanation for 
this could be that this series exhibits a single change point (according to 
most annotators, see Figure~\ref{fig:nile}), which also appears to be a clear 
shift in the mean of the series. We can also look at which time series 
appeared to be ``difficult'', in the sense that the methods failed to do well 
on it. Here, one noticeable case is the \verb+businv+ series where annotators 
generally agree on three change point locations, yet even in the Oracle 
experiment the maximum obtained F1 score is only 0.650.

Finally, we observe that the choice of evaluation metric affects the ranking 
of the algorithms (see Figure~\ref{fig:cd_diagrams}). Because the annotations 
are combined in a single set of ground truth change points when computing the 
precision, it is easier to achieve a high F1-score than it is to obtain a high 
score on the covering metric. Indeed, a high score on the covering metric 
indicates a high degree of agreement with each \emph{individual} annotator, 
whereas a high value on the F1-score indicates a strong agreement with the 
\emph{combined} set of annotations. Nevertheless, the best performing method 
is the same regardless of the metric, which reflects a degree of stability in 
the relative performance of the methods.

\section{Discussion}%
\label{sec:discussion}

We have introduced the first dedicated data set of human-annotated real-world 
time series from diverse application domains that is specifically designed for 
the evaluation and development of change point detection algorithms. Moreover, 
we have presented a framework for evaluating change point detection algorithms 
against ground truth provided by multiple human annotators, with two change 
point metrics that take distinct but complementary approaches to evaluating 
detection performance. With this data set we have conducted a thorough 
evaluation of a large selection of existing change point algorithms in an 
effort to discover those that perform best in practice. This benchmark study 
has shown that the binary segmentation algorithm of \citet{scott1974cluster} 
performs best on univariate time series when using the default algorithm 
parameters from \citet{killick2014changepoint}, but its performance is not 
statistically significantly different from several other change point 
detection methods. When hyperparameter tuning is performed, the \textsc{bocpd} 
method of \citet{adams2007bayesian} outperforms all other methods on both 
univariate and multivariate time series. However, the differences between it 
and some other algorithms (\textsc{binseg}, \textsc{segneigh}, and 
\textsc{rfpop}, among others) have been shown to not be statistically 
significantly different.

Both the introduced data set and the benchmark study can be extended and 
improved in the future, and our open-source code repository linked previously 
is set up to facilitate this. However, the present paper has already provided 
novel insights in the practical performance of change point detection 
algorithms and indicated potential topics for further research on algorithm 
development. For instance, future work could focus on incorporating automated 
hyperparameter tuning methods to bridge the gap between the Default and Oracle 
performance of the methods (perhaps taking inspiration from the work of  
\citealp{turner2009adaptive} for \textsc{bocpd}). It would also be relevant to 
understand why \textsc{binseg} performs as well or better than some of the 
more advanced methods, especially in the absence of hyperparameter tuning.  
Importantly, the presented change point data set and benchmark study motivate 
these research directions based on a quantitative analysis of algorithm 
performance, which was not previously possible.

Because of the unsupervised nature of the change point detection problem, it 
is difficult to compare algorithms in settings other than through simulation 
studies.  But since the goal is ultimately to apply these algorithms to 
real-world time series, we should strive to ensure that they work well in 
practice.  The change point data set presented here was inspired by the 
Berkeley Segmentation Data Set introduced in \citet{martin2001database} as 
well as the PASCAL VOC challenge \citep{everingham2010pascal}. In the past, 
these data sets have been crucial in accelerating the state of the art in image 
segmentation and object recognition, respectively. Our hope is that the 
data set and the benchmark study presented in this work can do the same for 
change point detection.

\section*{Acknowledgments}
The authors would like to thank James Geddes as well as the other annotators 
who preferred to remain anonymous. We are also grateful to Siamak Zamani 
Dadaneh for helpful comments on an earlier version of this manuscript, as well 
as for suggesting the \textsc{zero} baseline method. Finally, we would like to 
thank the anonymous reviewers for their comments, which have helped improve 
the paper. This work was supported in part by The Alan Turing Institute under 
EPSRC grant EP/N510129/1.

\section*{Changelog}
Below is a list of changes to the arXiv version of this paper.

\begin{itemize}
	\item v3
		\begin{itemize}
			\item Expanded the grid search for some of the methods 
				in the Oracle experiment
			\item Changed from rank plots to critical-difference 
				diagrams
			\item Updated the results section taking into account 
				the above changes
			\item Expanded analysis of annotator agreement
			\item Various minor corrections and clarifications
		\end{itemize}
	\item v2
		\begin{itemize}
			\item Added the \textsc{zero} method to the 
				comparison.
			\item Added summary statistics in the text regarding 
				the length of the time series and the number 
				of annotated change points.
			\item Added rank plots for multivariate time series to 
				the appendix.
			\item Corrected an error in the computation of the F1 
				score and updated the results and the online 
				code repository. This correction had no major 
				effect on our conclusions.
			\item Made minor revisions in 
				Section~\ref{sub:results} and 
				Section~\ref{sec:discussion} to reflect minor 
				changes in the relative performance of the 
				algorithms.
			\item Updated acknowledgements.
		\end{itemize}
	\item v1
		\begin{itemize}
			\item Initial version.
		\end{itemize}
\end{itemize}

\clearpage
\appendix

\section{Simulation Details}%
\label{app:simulation_details}

This section presents additional details of the implementations and 
configurations used for each of the algorithms. Where possible, we used 
packages made available by the authors of the various methods. During the grid 
search those hyperparameter configurations that led to computational errors 
(e.g.,~numerical overflow) were skipped. While we describe our experimental 
setup in detail below, we also recommend the reader to check our online 
repository to reproduce our setup exactly.\footnote{See: 
	\url{https://github.com/alan-turing-institute/TCPDBench}.}

\begin{itemize}
	\item \textsc{amoc}, \textsc{binseg}, \textsc{pelt}, and 
		\textsc{segneigh} were implemented using the R 
		\verb+changepoint+ package, v2.2.2  
		\citep{killick2014changepoint}.
		\begin{itemize}
			\item As default parameters we used the 
				\verb+cpt.mean+ function, following 
				\citet{fryzlewicz2014wild}, with the Modified 
				Bayes Information Criterion (MBIC) penalty, 
				the Normal test statistic, and minimum segment 
				length of 1. For \textsc{binseg} and 
				\textsc{segneigh} we additionally used the 
				default value of $5$ for the maximum number of 
				change points/segments.
			\item During the grid search, we varied the parameters 
				as follows:
				\begin{description}
					\item [Function:] \verb+cpt.mean+, 
						\verb+cpt.var+, and
						\verb+cpt.meanvar+.
					\item [Penalty:] None, SIC, BIC, MBIC, 
						AIC, Hannan-Quinn, Asymptotic 
						(using p-value 0.05), and 
						Manual (with a grid of 101 
						values for the regularization 
						parameter, evenly spaced on a 
						log scale between $10^{-3}$ 
						and $10^{3}$).
					\item [Test Statistic:] Normal, CUSUM, 
						CSS, Gamma, Exponential, and 
						Poisson (where appropriate for 
						the specified function).
					\item [Max. CP:] 5 (default) or $T/2 + 
						1$ (max). Only for 
						\textsc{binseg} or 
						\textsc{segneigh}.
				\end{description}
				Combinations of parameters that were invalid 
				for the method were ignored.
		\end{itemize}
	\item \textsc{cpnp} was implemented with the R \verb+changepoint.np+ 
		package (v0.0.2). The penalty was varied in the same way as 
		for \textsc{pelt} (see above) with the addition that in the 
		grid search the number of quantiles was varied on the grid 
		$\{10, 20, 30, 40\}$.
	\item \textsc{rfpop} was called using the \verb+robseg+ package 
		available on 
		GitHub.\footnote{\url{https://github.com/guillemr/robust-fpop}, 
			version 2019-07-02.} As default setting the biweight 
		loss function was used. During the grid search the parameters 
		were varied as follows:
		\begin{description}
			\item[Loss:] \verb+L1+, \verb+L2+, \verb+Outlier+, 
				\verb+Huber+
			\item[Penalty value:] \verb+lambda+ was varied on a 
				grid of 101 values evenly spaced on a log 
				scale between $10^{-3}$ and $10^{3}$ (as above 
				for the methods in the \verb+changepoint+ 
				package).
			\item[Threshold:] the \verb+lthreshold+ parameter was 
				varied on a grid of 11 values evenly spaced on 
				a log scale between $10^{-1}$ and $10^1$.
		\end{description}
		We observe that because in addition to the loss function this 
		method has two parameters that need to be evaluated on a grid, 
		it has the largest number of hyperparameter configurations of 
		any method in the Oracle experiment ($2,424$ in total).
	\item \textsc{ecp} and \textsc{kcpa} were accessed through the R
		\verb+ecp+ package (v3.1.1) available on CRAN 
		\citep{james2013ecp}.  For \textsc{ecp} the parameter settings 
		were:
		\begin{itemize}
			\item As default parameters we used the 
				\verb+e.divisive+ function with $\alpha = 1$, 
				minimum segment size of 30, 199 runs for the 
				permutation test and a significance level of 
				0.05.
			\item The grid search consisted of:
				\begin{description}
					\item [Algorithm:] \verb+e.divisive+ 
						or \verb+e.agglo+
					\item [Significance level:] 0.05 or 
						0.01.
					\item [Minimum segment size:] 2 
						(minimal) or 30 (default)
					\item [$\alpha$]: $(0.5, 1.0, 1.5)$
				\end{description}
				For the \verb+e.agglo+ algorithm only the 
				$\alpha$ parameter has an effect.
		\end{itemize}
		The \textsc{kcpa} method takes two parameters: a cost and a 
		maximum number of change points. In the default experiment the 
		cost parameter was set to 1 and we allowed the maximum number 
		of change points possible. In the grid search the cost 
		parameter was varied in the same way as above for the methods 
		in the \verb+changepoint+ package, on a grid of 101 values 
		evenly spaced on a log scale between $10^{-3}$ and $10^3$,
		and the maximum number of change points was varied between the 
		maximum possible and $5$, which corresponds to the methods in 
		the \verb+changepoint+ package.
	\item \textsc{wbs} was evaluated using the R package of the same name 
		available on CRAN (v1.4).
		\begin{itemize}
			\item As defaults we used the SSIC penalty, the 
				integrated version, and the default of 50 for 
				the maximum number of change points.
			\item During the grid search we used:
				\begin{description}
					\item [Penalty:] SSIC, BIC, MBIC
					\item [Integrated:] true, false
					\item [Max CP:] 50 (default) or $T$ 
						(max)
				\end{description}
		\end{itemize}
	\item \textsc{bocpd} was called using the R package \verb+ocp+ 
		(v0.1.1), available on CRAN \citep{pagotto2019ocp}.  For the 
		experiments we used the Gaussian distribution with the 
		negative inverse gamma prior.  The prior mean was set to $0$, 
		since the data is standardized.  No truncation of the run 
		lengths was applied.
		\begin{itemize}
			\item For the default experiment the intensity 
				($\lambda_{\text{gap}}$) was set to 100, the 
				prior variables were set to $\alpha_0 = 1$, 
				$\beta_0 = 1$, and $\kappa_0 = 1$.
			\item For the grid search, we varied the parameters as 
				follows:
				\begin{description}
					\item [Intensity:] $(10, 50, 100, 200)$
					\item [$\alpha_0$:] $(0.01, 0.1, 1, 10, 100)$
					\item [$\beta_0$:] $(0.01, 0.1, 1, 10, 100)$
					\item [$\kappa_0$:] $(0.01, 0.1, 1, 10, 100)$
				\end{description}
		\end{itemize}
	\item \textsc{bocpdms} and \textsc{rbocpdms} were run with the code 
		available on 
		GitHub.\footnote{\url{https://github.com/alan-turing-institute/bocpdms/} 
			at Git commit hash f2dbdb5, and 
			\url{https://github.com/alan-turing-institute/rbocpdms/} 
			at Git commit hash 643414b, respectively.} The 
		settings for \textsc{bocpdms} were the same as that for 
		\textsc{bocpd} above, with the exception that during the grid 
		search run length pruning was applied by keeping the best 100 
		run lengths. This was done for speed considerations. No run 
		length pruning was needed for \textsc{bocpdms} with the 
		default configuration, but this was applied for 
		\textsc{rbocpdms}.  For \textsc{rbocpdms} we used the same 
		settings, but supplied additional configurations for the 
		$\alpha_0$ and $\alpha_{\text{rld}}$  parameters.  As default 
		values and during the grid search we used $\alpha_0 = 
		\alpha_{\text{rld}} = 0.5$.  For both methods we applied a 
		timeout of 30 minutes during the grid search and we used a 
		timeout of 4 hours for \textsc{rbocpdms} in the default 
		experiment.
	\item \textsc{prophet} was run in R using the \verb+prophet+ package 
		(v0.4). As Prophet requires the input time vector to be a 
		vector of dates or datetimes, we supplied this where available 
		and considered the series that did not have date information 
		to be a daily series. By supplying the dates of the 
		observations Prophet automatically chooses whether to include 
		various seasonality terms. For both experiments we allowed 
		detecting of change points on the entire input range and used 
		a threshold of $0.01$ for selecting the change points (in line 
		with the change point plot function in the Prophet package).  
		During the grid search we only varied the maximum number of 
		change points between the default (25) and the theoretical 
		maximum ($T$).
\end{itemize}

\section{Simulated Annotator Agreement}%
\label{app:annotator_agreement}

In this section we present the results of a simulation study that aims to 
quantify the extent to which the annotator agreement described in 
Section~\ref{sub:consistency} can be expected by chance.

For every time series in the data set we simulate a set of change point 
annotations for each of the $K = 5$ annotators. Next, for each annotator we 
compute the one-vs-rest (OvR) annotator agreement compared to the simulated 
other annotators, as described in the main text. These values are then 
averaged into a single value that represents the simulated average OvR 
annotator agreement. By repeating this process a large number of times we 
obtain a distribution over the average OvR annotator agreement for a 
particular time series. Naturally, we can also compute the average OvR 
annotator agreement for the human annotators. We can then compute the 
$p$-value for the hypothesis that the average OvR annotator agreement for the 
human annotators is larger than what is expected by chance. Formally, if $X$ 
is a random variable reflecting the distribution of the simulated average OvR 
annotator agreement and $\bar{x}$ is the observed human OvR annotator 
agreement, then we are interested in the probability $p(X \geq \bar{x})$.


Recall from the main text that $n_k$ is the number of change points identified 
by annotator $k$, and $\tau_i$ is the index of change point $i \in \{1,\ldots, 
n_k\}$. We simulate the annotations as follows for each simulated annotator $k 
\in \{1,\ldots, K\}$,
\begin{align}
	n_k &\sim \text{Pois}(\eta)\\
	\tau_i &\sim \text{Unif}(2, T - 1),\qquad \forall i \in \{1, \ldots, 
	n_k\},
\end{align}
where $\eta$ is a hyperparameter for the simulation discussed below.
Note that the domain for the change points $\tau_i$ is chosen such that no 
change points can be declared at the first and last time index.\footnote{For 
	very short time series (e.g.,~{\tt centralia}) the value of $n_k$ may 
	exceed the length of the series $T$ if the rate of the Poisson 
	distribution, $\eta$, is set too high. If that occurs in the 
	simulation $n_k$ is capped at the maximum possible value of $T - 2$.}  
For each time series in the data set we simulate 100,000 sets of annotations 
and compute the average OvR annotator agreement. The $p$-value for each series 
is then approximated as the proportion of realizations of $X$ that are larger 
or equal to $\bar{x}$. The rate of the Poisson distribution, $\eta$, affects 
the number of change points for each simulated annotator. A priori, there is 
no information on how this rate should be chosen such that the simulation is 
as close as possible to the real-world case. Thus, we propose to set $\eta$ to 
the number of change points declared per human annotator averaged over all 
time series in the data set ($\eta \approx 2.295$).

\begin{table}[tb]
	\centering
	\scriptsize
	\input{./analysis_output/tables/annotation_simulation_pvalues.tex}
	\caption{Simulated $p$-values for the average one-vs-rest annotator 
		agreement observed for the human annotators. Values larger 
		than a significance level of $0.05$ are highlighted in bold.
		\label{tab:pvalues}}
\end{table}

Using the above simulation setup we obtain the $p$-values listed in 
Table~\ref{tab:pvalues}. We observe that conditional on the assumptions made 
in the simulation, for most time series in the data set the observed average 
OvR annotator agreement for the human annotators is significantly larger than 
what can be expected by chance.  For the F1-score we find a $p$-value larger 
than $0.05$ only for the \verb+centralia+ time series.  This is unsurprising 
for two reasons: this series consists of only 15 observations, and there is 
disagreement amongst the annotators on the existence of change points in the 
series. For the covering metric there are few additional series where the 
$p$-value obtained from the simulation is larger than $0.05$. This can be 
explained by the fact that this metric rewards high agreement with each 
\emph{individual} annotator, and in these series there is some amount of 
disagreement among the annotators.

\FloatBarrier
\section{Additional Tables and Figures}
\label{app:additional_tables_figures}

Below we present additional tables and figures to support the material in the 
main text. In some tables the results on the quality control time series are 
included for completeness, but these are not used in the benchmark study.

\subsection{Tables}%
\label{app:tables}

\begin{table}[ht]
	\scriptsize
	\centering
	\caption{Descriptive statistics of the time series included in the 
		data set. The three rightmost columns respectively show the 
		minimum, maximum, and average number of change points marked 
		by the annotators (see Section~\ref{app:datasets} for the 
		complete annotations). The series marked with a dagger 
		($\dagger$) are financial time series that only yield 
		observations on trading days, which were additionally sampled 
		every three days to reduce their length.  
		\label{tab:descriptive}}
	\input{./analysis_output/tables/descriptive_statistics.tex}
\end{table}

\begin{table}[ht]
	\tiny
	\centering
	\caption{\textbf{Default-Covering.} Results for each method on each of 
		the time series for the experiment with default settings using 
		the covering metric. Values are unavailable either due to a 
		failure (F), because of missing values in the series 
		unsupported by the method (M), or because the method is not 
		suited for multidimensional series (blank). Highest values for 
		each time series are shown in bold.}
	\hspace*{-20mm}\input{./analysis_output/tables/default_cover_combined_full.tex}
\end{table}

\begin{table}[ht]
	\tiny
	\centering
	\caption{\textbf{Default-F1.} Results for each method on each of the 
		time series for the experiment with default settings using the 
		F1-score. Values are unavailable either due to a failure (F), 
		because of missing values in the series unsupported by the 
		method (M), or because the method is not suited for 
		multidimensional series (blank).  Highest values for each time 
		series are shown in bold.}
	\hspace*{-2cm}\input{./analysis_output/tables//default_f1_combined_full.tex}
\end{table}

\begin{table}[ht]
	\tiny
	\centering
	\caption{\textbf{Oracle-Covering.} Maximum segmentation covering score 
		for each method on each of the time series after running the 
		grid search. Values are unavailable either due to a failure 
		(F), because of missing values in the series unsupported by 
		the method (M), the method timing out (T), or because the 
		method is not suited for multidimensional series (blank).  
		Highest values for each time series are shown in bold.}
	\hspace*{-2cm}\input{./analysis_output/tables/oracle_cover_combined_full.tex}
\end{table}

\begin{table}[ht]
	\tiny
	\centering
	\caption{\textbf{Oracle-F1.} Maximum F1-score for each method on each 
		of the time series after running the grid search.  Values are 
		unavailable either due to a failure (F), because of missing 
		values in the series unsupported by the method (M), the method 
		timing out (T), or because the method is not suited for 
		multidimensional series (blank).  Highest values for each time 
		series are shown in bold.}
	\hspace*{-2cm}\input{./analysis_output/tables/oracle_f1_combined_full.tex}
\end{table}

\FloatBarrier

\subsection{Multivariate Critical Difference Diagrams}%
\label{app:multivariate_cd_diagrams}

\begin{figure*}[ht]
	\def\RankFigWidth{0.47}
	\centering
	\subfloat[][Default -- covering metric]{%
		\centering
		\includegraphics[width=\RankFigWidth\linewidth]{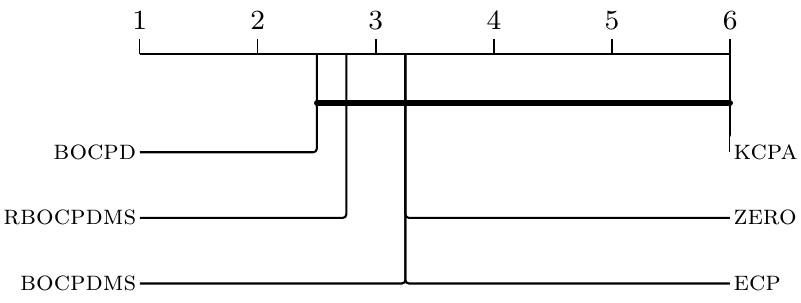}
		\label{fig:default_cd_cover_multi}
	}%
	\quad
	\subfloat[][Default -- F1-score]{%
		\centering
		\includegraphics[width=\RankFigWidth\linewidth]{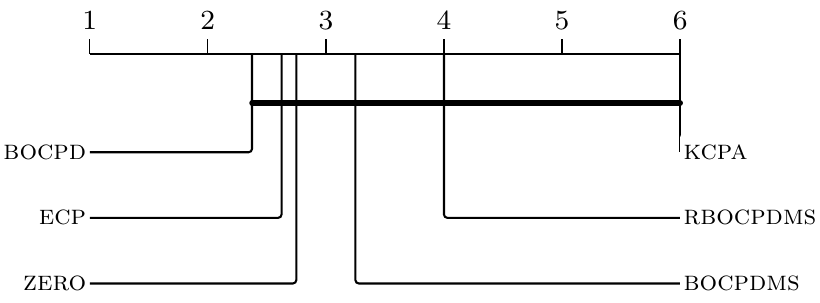}
		\label{fig:default_cd_f1_multi}
	}%
	\\
	\subfloat[][Oracle -- covering metric]{%
		\centering
		\includegraphics[width=\RankFigWidth\linewidth]{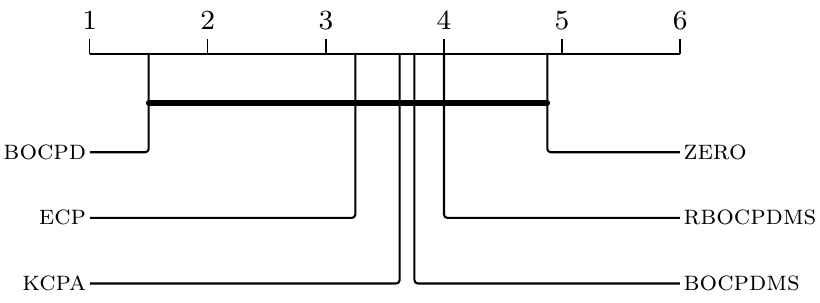}
		\label{fig:oracle_cd_cover_multi}
	}%
	\quad
	\subfloat[][Oracle -- F1-score]{%
		\centering
		\includegraphics[width=\RankFigWidth\linewidth]{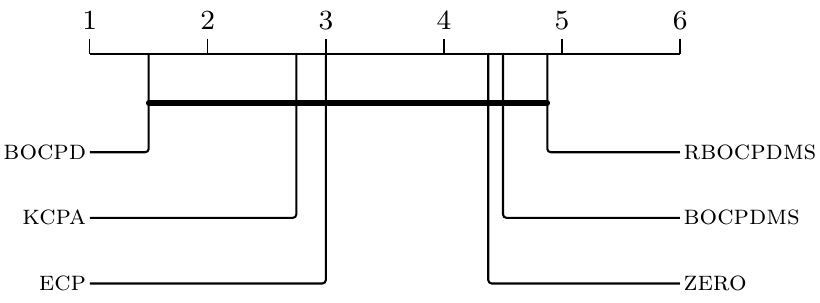}
		\label{fig:oracle_cd_f1_multi}
	}
	\caption{Critical difference diagrams for both experiments and both 
		metrics for the multivariate time series. Methods that are not 
		significantly different (at $\alpha~=~0.05$) are connected.  
		Lower ranks are better.}
	\label{fig:rankplots_multi}
\end{figure*}

\FloatBarrier

\section{Data Set Overview}
\label{app:datasets}

In the following pages we show the time series included in the change point 
data set, as well as the provided annotations. Annotations are marked on the 
graphs by a dashed vertical line with one or more triangles on the horizontal 
axis that correspond to the number of annotators that marked that point as a 
change point. The colors of the triangles correspond to each of the five 
annotators for the time series. The number in the box on the left side of the 
figure is the number of annotators who believed there were no change points in 
the time series. A brief description including the source of the series is 
provided.\footnote{See \url{https://github.com/alan-turing-institute/TCPD} for 
	the complete data set and further information on each individual time 
	series.}

\subsection{Real Time Series}%
\label{sub:real_time_seris}

This section lists the real time series from various sources and application 
domains.

\MyFigure{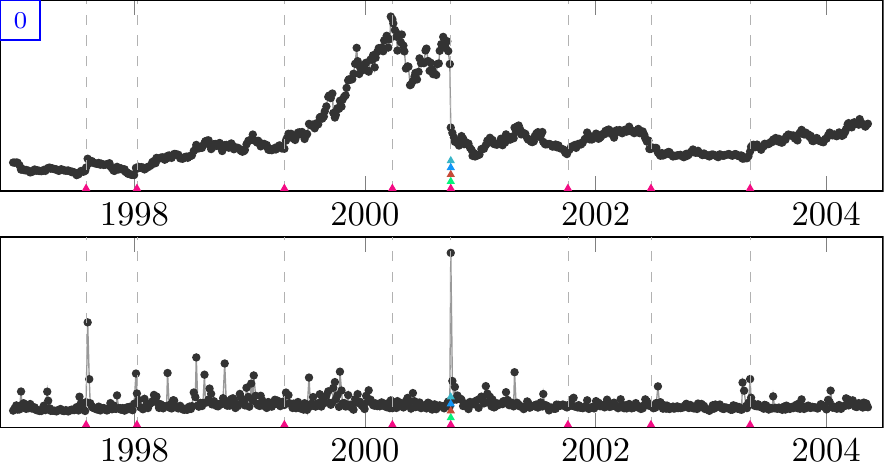}{apple}{apple}{The daily closing price and volume of Apple, 
	Inc. stock for a period around the year 2000. The series was sampled 
	at every three time steps to reduce the length of the series. A 
	significant drop in stock price occurs on 2000-09-29. Data retrieved 
	from Yahoo Finance.}
\MyFigure{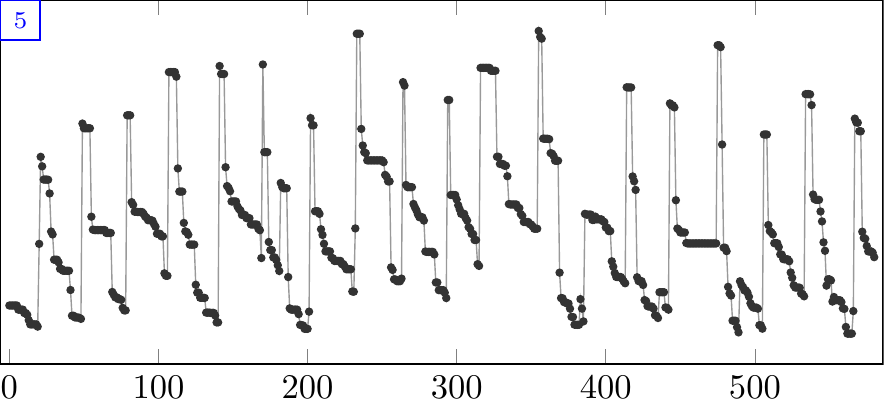}{bank}{bank}{The amount of money in someone's current account.  
	Significant changes occur, but the series is also periodic.}
\MyFigure{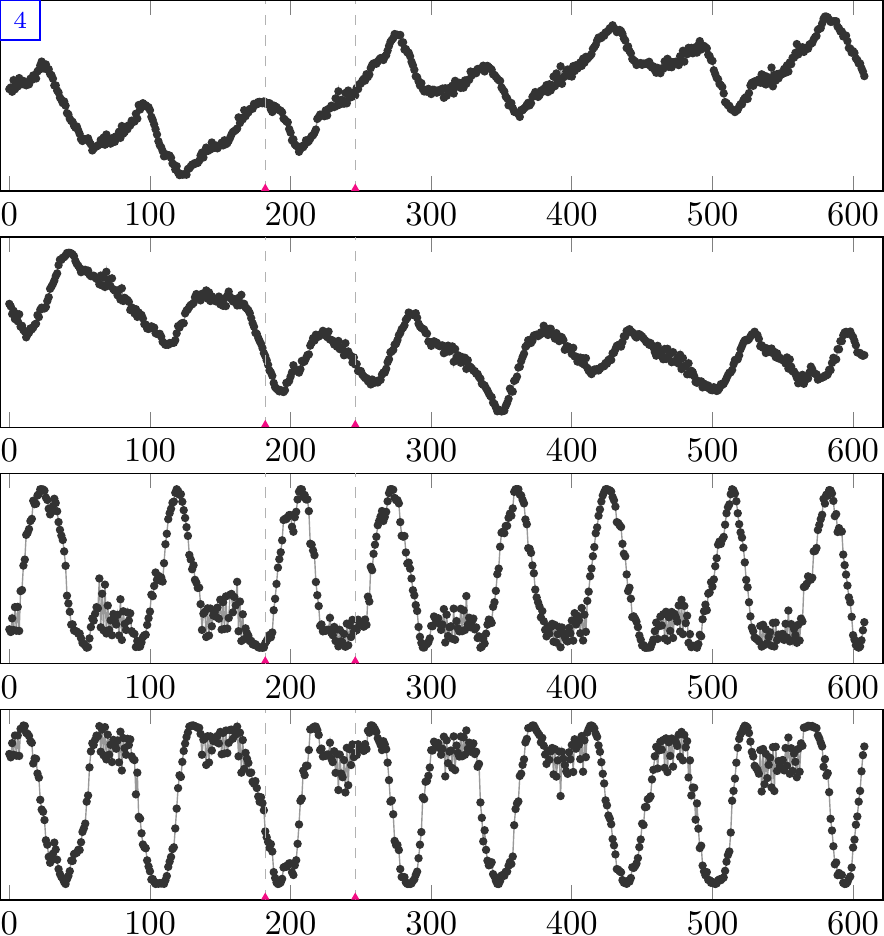}{beewaggle6}{bee\_waggle\_6}{Honey bee movement 
	switches between three states: left turn, right turn, and waggle. This 
	series contains the $x$ position, $y$ position, and the sine and 
	cosine of the head angle of a single bee. Data obtained from 
	\citet{oh2008learning}.}
\MyFigure{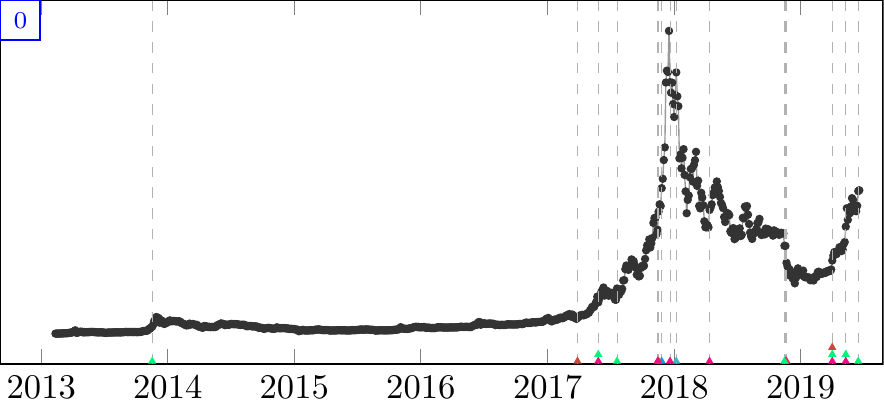}{bitcoin}{bitcoin}{The price of Bitcoin in USD. Data 
	obtained from Blockchain.com.}
\MyFigure{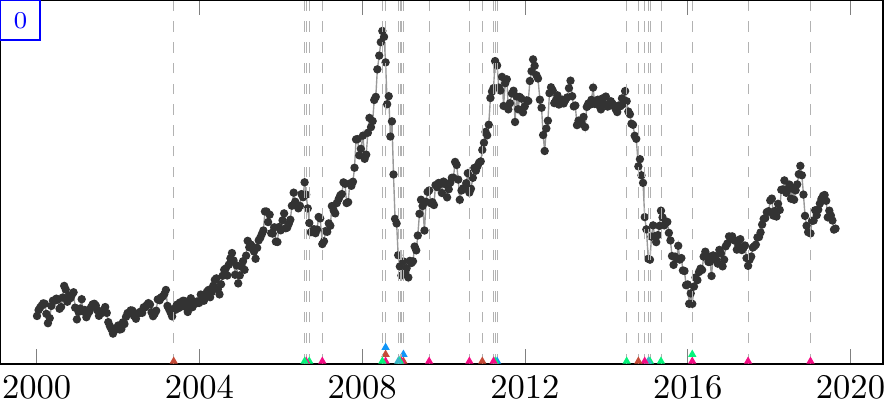}{brentspot}{brent\_spot}{The price of Brent Crude oil in 
	USD per barrel. Significant changes appear around 2008 and 2014.  The 
	data is sampled at every 10 original observations to reduce the length 
	of the series.  Obtained from the U.S.~Energy Information 
	Administration.}
\MyFigure{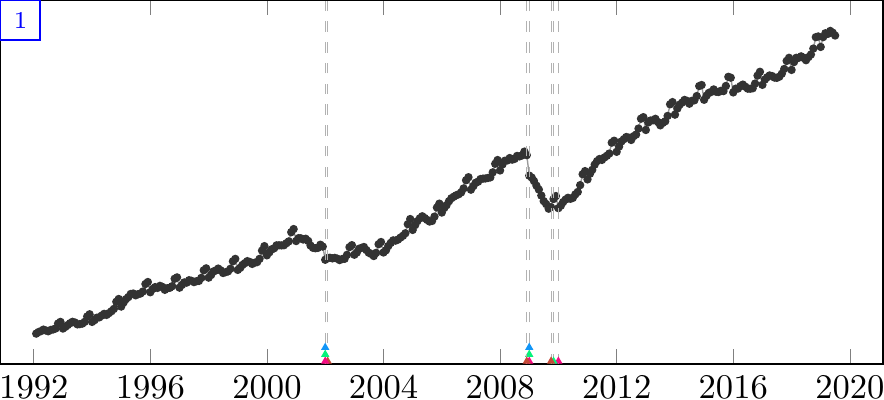}{businv}{businv}{Monthly total business inventories (in USD), 
	obtained from the U.S.~Census Bureau. Effects of the financial crisis 
	are visible. The non-seasonally-adjusted series is used to maintain 
	the periodic component.}
\MyFigure{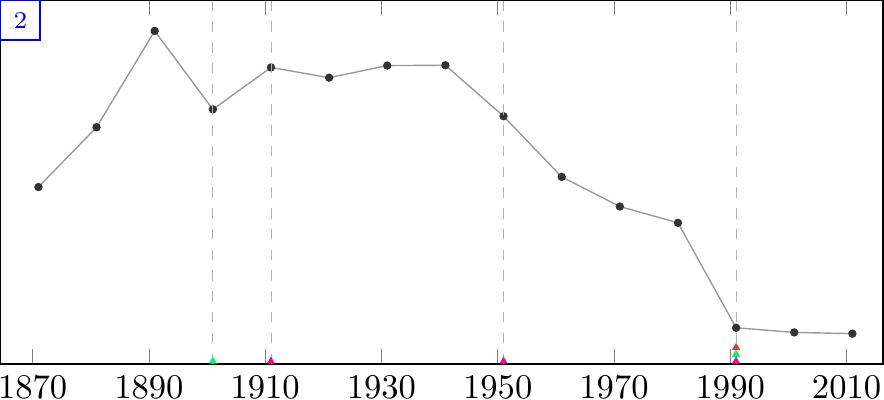}{centralia}{centralia}{The population of the mining town 
	of Centralia, Pennsylvania, U.S., where a mine fire has been burning 
	since 1962.  Eminent domain was invoked in 1992, condemning all 
	buildings in the town. Data obtained from Wikipedia.}
\MyFigure{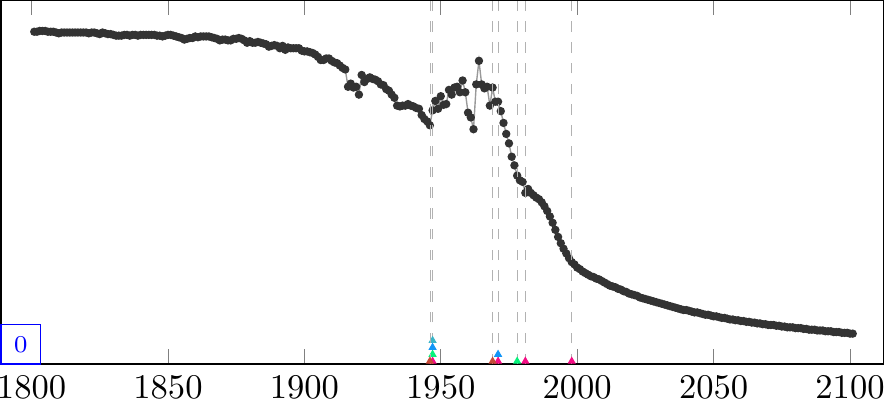}{childrenperwoman}{children\_per\_woman}{The 
	global average number of children born per woman. Data obtained from 
	GapMinder.}
\MyFigure{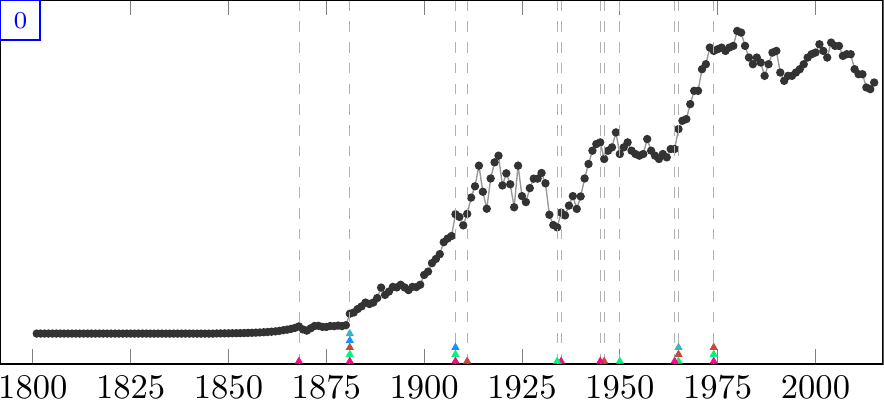}{co2canada}{co2\_canada}{$\text{CO}_2$ emissions per 
	person in Canada.  Data obtained from GapMinder.}
\MyFigure{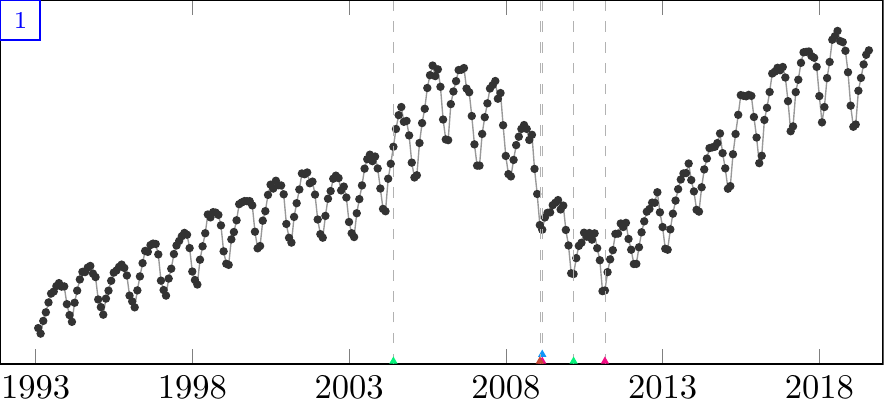}{construction}{construction}{Total private construction 
	spending in the U.S. Data obtained from the U.S.~Census Bureau.}
\MyFigure{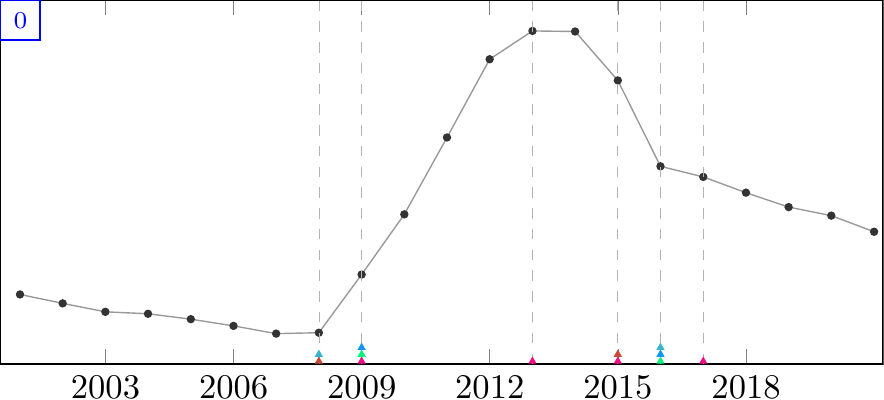}{debtireland}{debt\_ireland}{The government debt ratio 
	of Ireland.  Effects of the financial crisis of 2007--2008 are 
	visible.  Data obtained from EuroStat.}
\MyFigure{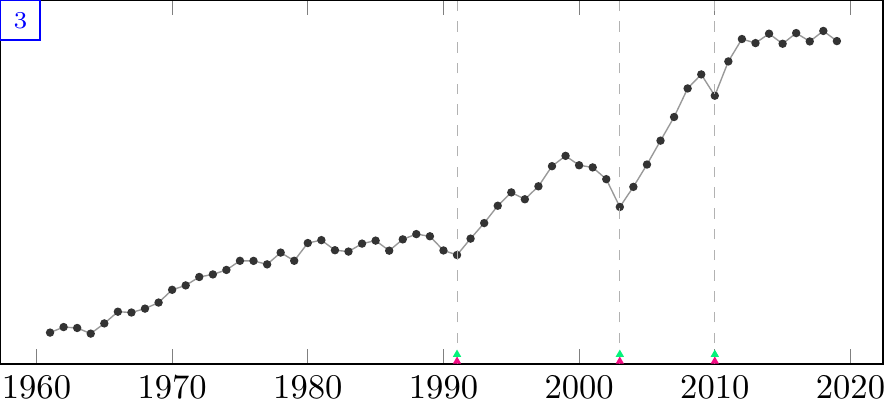}{gdpargentina}{gdp\_argentina}{The GDP of Argentina in 
	constant local currency. Data obtained from the World Bank.}
\MyFigure{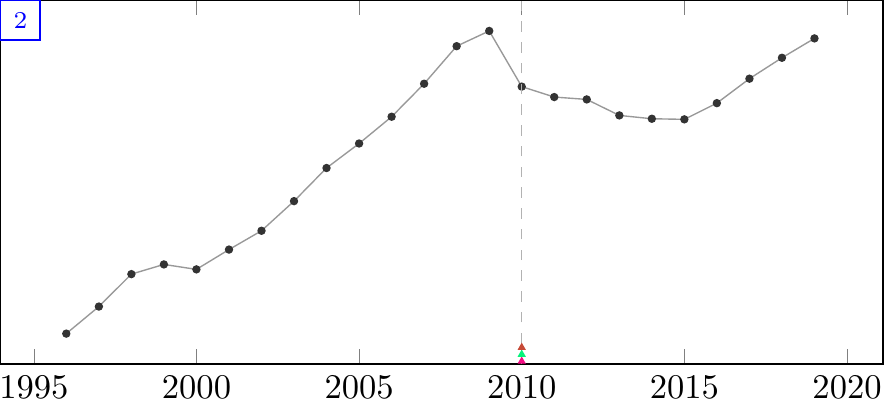}{gdpcroatia}{gdp\_croatia}{The GDP of Croatia in 
	constant local currency. Data obtained from the World Bank.}
\MyFigure{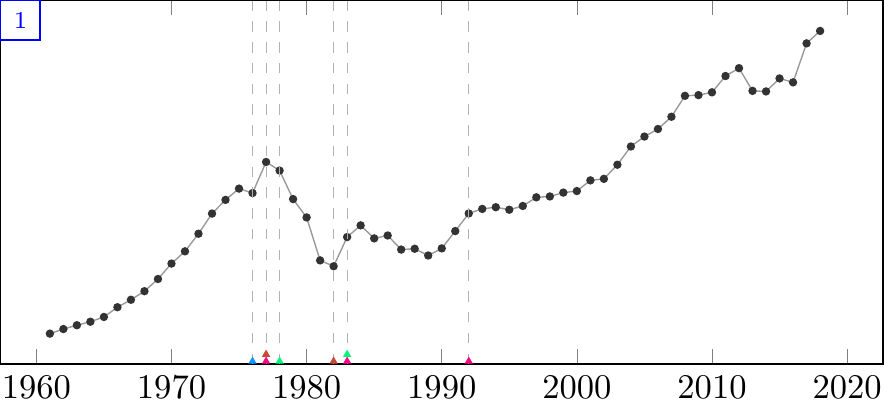}{gdpiran}{gdp\_iran}{The GDP of Iran in constant local 
	currency. Data obtained from the World Bank.}
\MyFigure{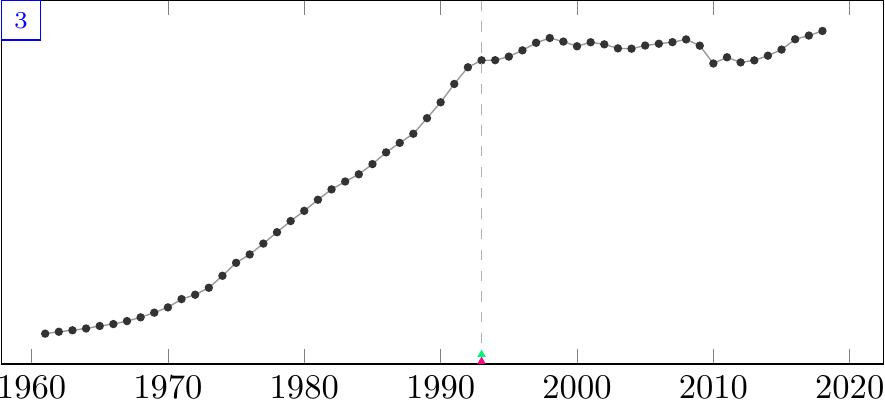}{gdpjapan}{gdp\_japan}{The GDP of Japan in constant local 
	currency. Data obtained from the World Bank.}
\MyFigure{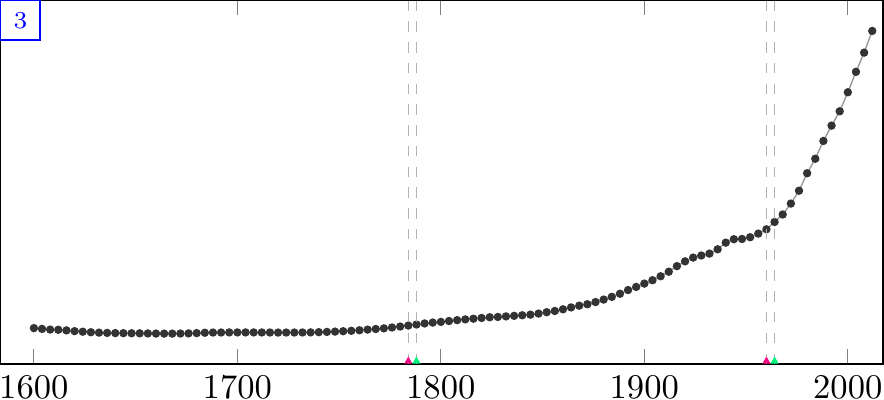}{globalco2}{global\_co2}{Monthly Global $\text{CO}_2$ 
	levels. The series has been sampled every 48 months and cropped to 
	recent history to reduce the length of the series. Data obtained from 
	\citet{meinshausen2017historical}.}
\MyFigure{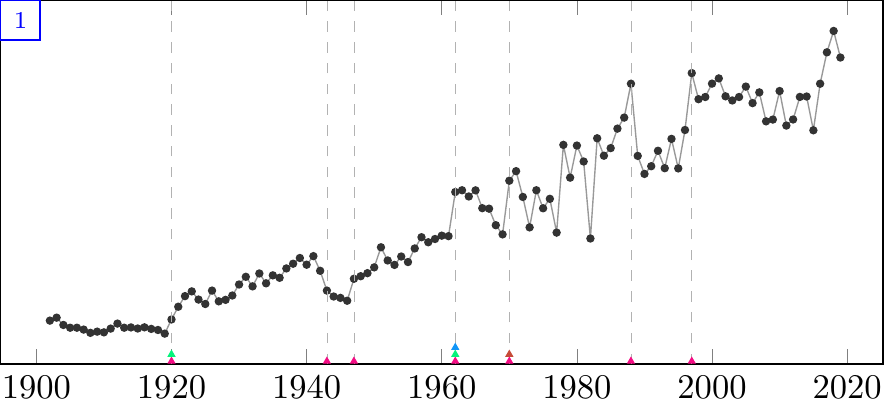}{homeruns}{homeruns}{The number of home runs in the 
	American League of baseball by year. Potentially significant events 
	are the second world war and the Major League Baseball expansions.  
	Data retrieved from the Baseball Databank.}
\MyFigure{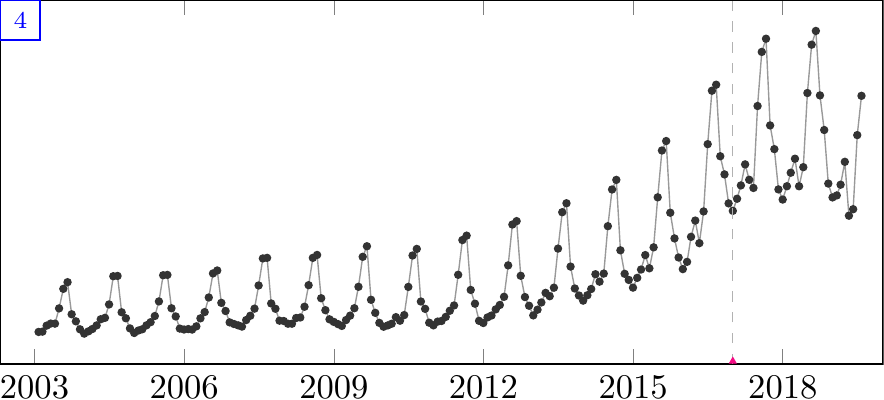}{iceland}{iceland\_tourism}{Monthly visitors to 
	Iceland through Keflavik airport. Data obtained from the Icelandic 
	Tourist Board.}
\MyFigure{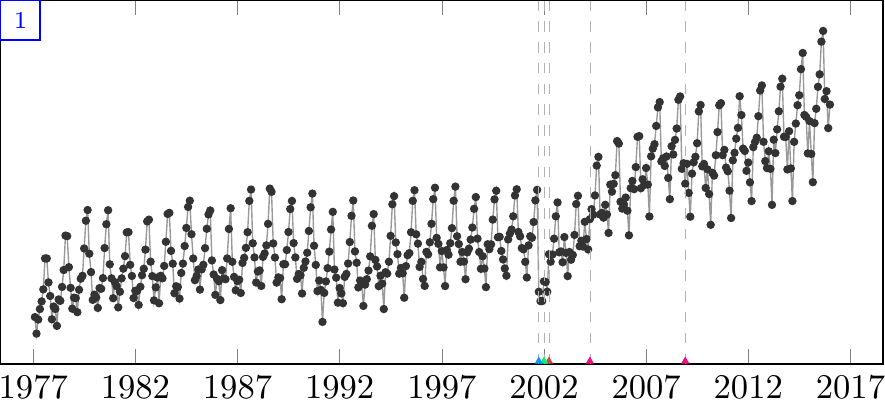}{jfkpassengers}{jfk\_passengers}{The number of 
	passengers arriving and departing at John F. Kennedy airport in New 
	York City.  Data obtained from the Port Authority of New York and New 
	Jersey.}
\MyFigure{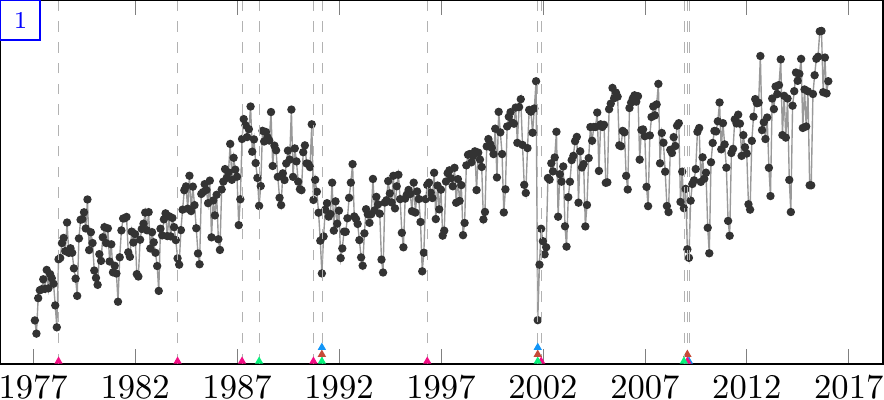}{lgapassengers}{lga\_passengers}{The number of 
	passengers arriving and departing at LaGuardia airport in New York 
	City. Data obtained from the Port Authority of New York and New 
	Jersey.}
\MyFigure{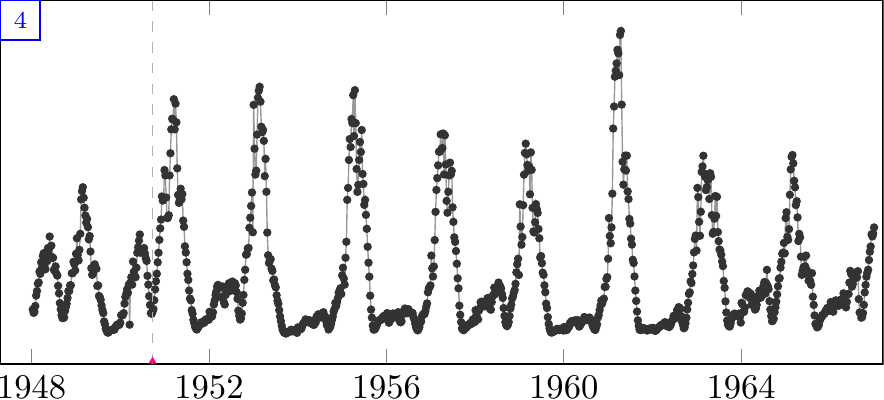}{measles}{measles}{Number of measles cases in England and 
	Wales over time. Data obtained from 
	\url{https://ms.mcmaster.ca/~bolker}.}
\MyFigure{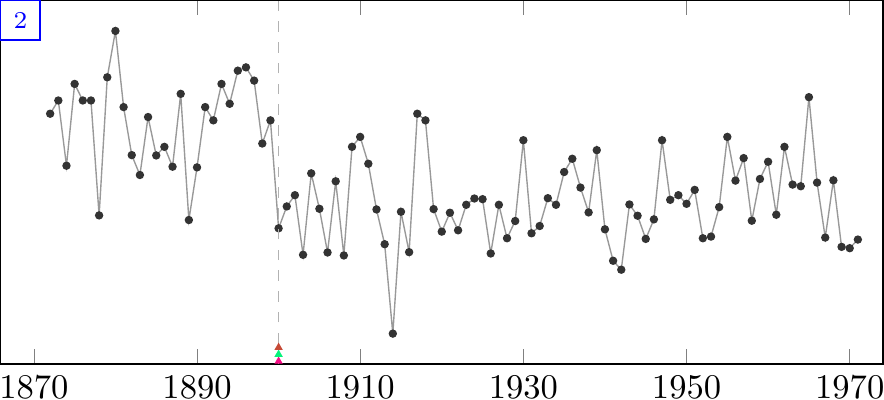}{nile}{nile}{Yearly volume of the Nile river at Aswan. A dam 
	was built in 1898. Data obtained from the website for the book by 
	\citet{durbin2012time}. This data set differs from the Nile data set 
	for the period 622 to 1284 AD, used in other change point work.}
\MyFigure{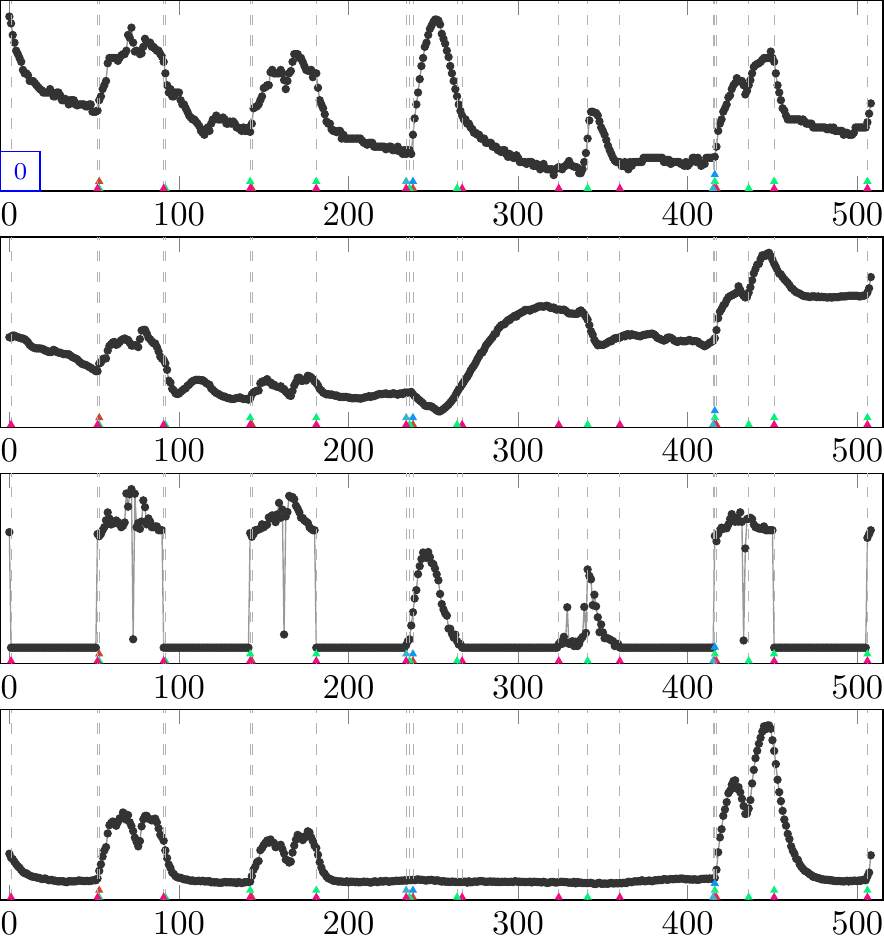}{occupancy}{occupancy}{Room occupancy time series by 
	\citet{candanedo2016accurate} obtained from the UCI repository 
	\citep{bache2013uci}. The four dimensions correspond to measurements 
	of temperature, relative humidity, light, and $\text{CO}_2$. The 
	original data has been sampled at every 16 observations to reduce the 
	length of the series.}
\MyFigure{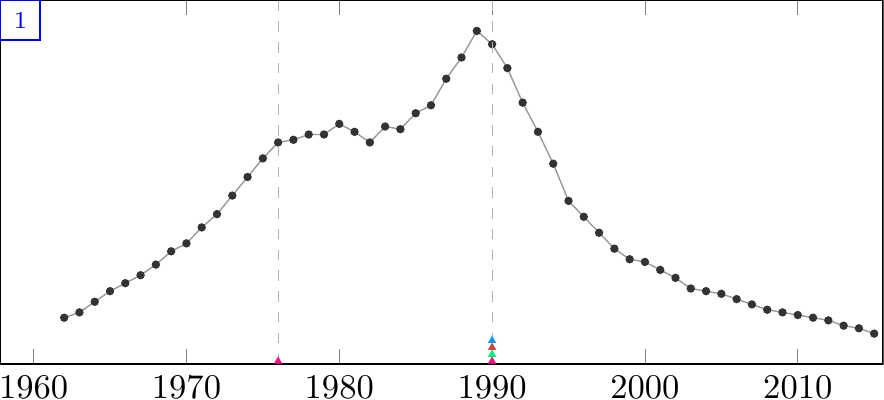}{ozone}{ozone}{Levels of ozone-depleting substances in the 
	atmosphere. The Montreal Protocol came into force in September 1989.  
	Data from \url{www.ourworldindata.org} and originally due to 
	\citet{hegglin2015twenty}.}
\MyFigure{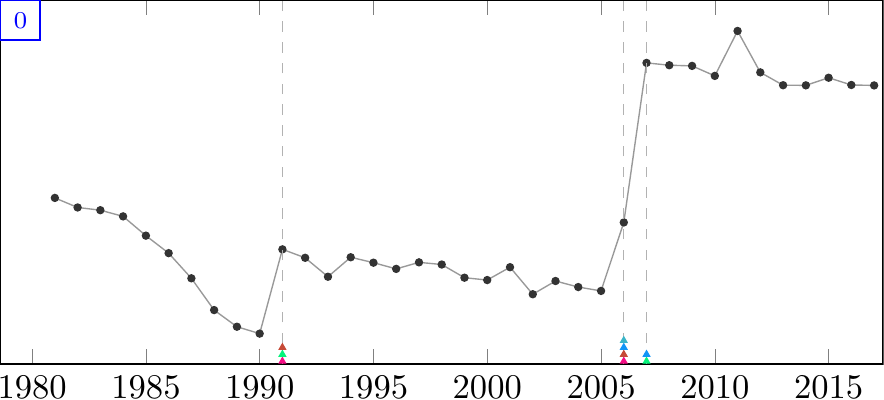}{raillines}{rail\_lines}{Total kilometers of rail lines 
	in the world.  Data obtained from the World Bank.}
\MyFigure{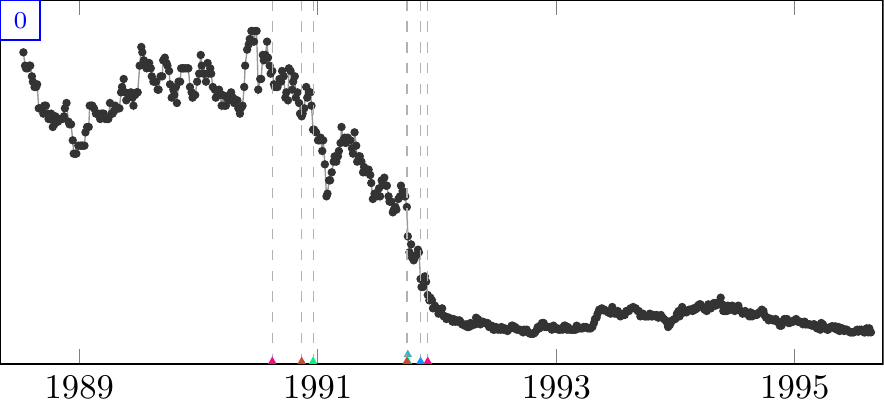}{ratnerstock}{ratner\_stock}{Stock price of the Ratner 
	Group for a period in the early 1990s. Data obtained from Yahoo 
	Finance and sampled every 3 observations to reduce the length of the 
	series.}
\MyFigure{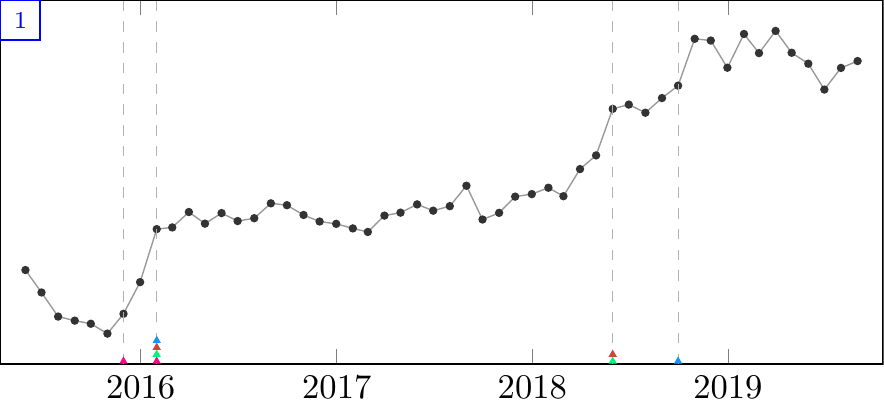}{robocalls}{robocalls}{The number of automated phone calls 
	in the U.S., which have been subject to varying degrees of regulation.  
	Obtained from \url{www.robocallindex.com}.}
\MyFigure{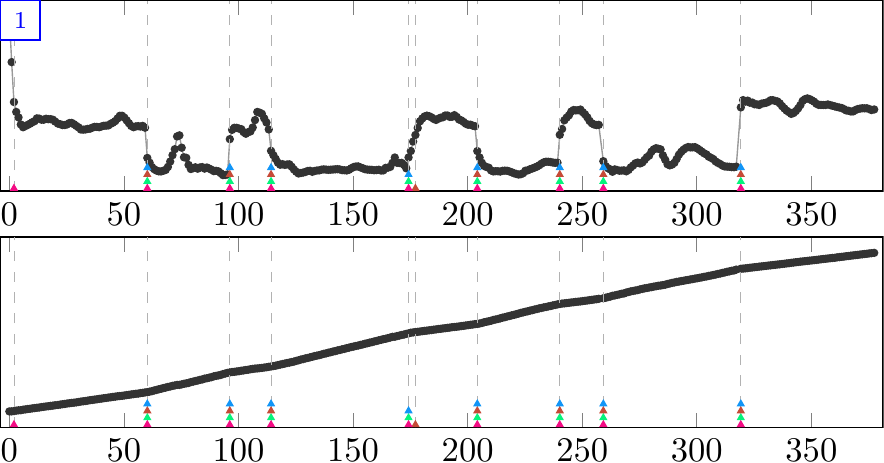}{runlog}{run\_log}{The pace and total distance traveled by a 
	runner following an interval training program (alternating between 
	running and walking). Data provided by the author.}
\MyFigure{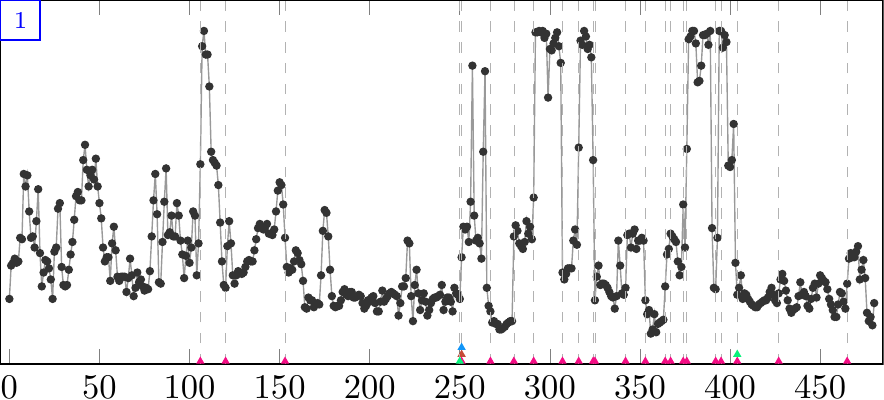}{scanline126007}{scanline\_126007}{A horizontal scan 
	line of image no. 126007 from the Berkeley Segmentation Data Set 
	\citep{martin2001database}.}
\MyFigure{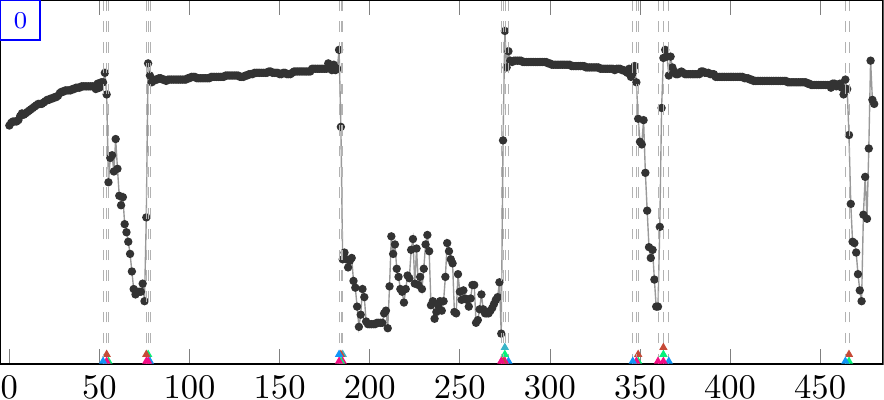}{scanline42049}{scanline\_42049}{A horizontal scan 
	line of image no. 42049 from the Berkeley Segmentation Data Set 
	\citep{martin2001database}.}
\MyFigure{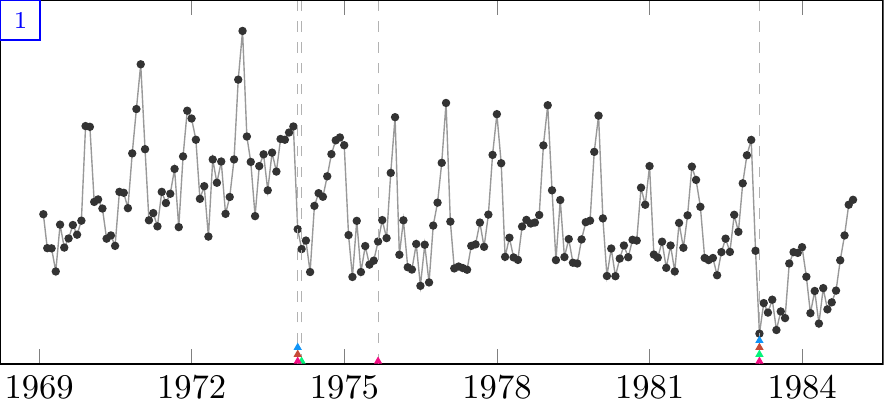}{seatbelts}{seatbelts}{Number of drivers killed or 
	seriously injured in the U.K.~around the period of the introduction of 
	seatbelts.  Seatbelts were compulsory in new cars starting in 1972 and 
	were mandatory to be worn from 1983 onwards. Obtained from the 
	\texttt{datasets} package in R.}
\MyFigure{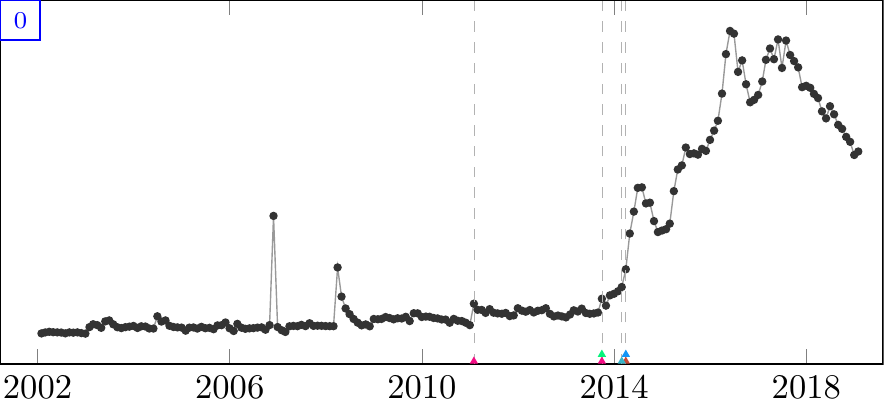}{shanghailicense}{shanghai\_license}{The number of 
	applicants for a license plate in Shanghai. Data obtained from 
	Kaggle.}
\MyFigure{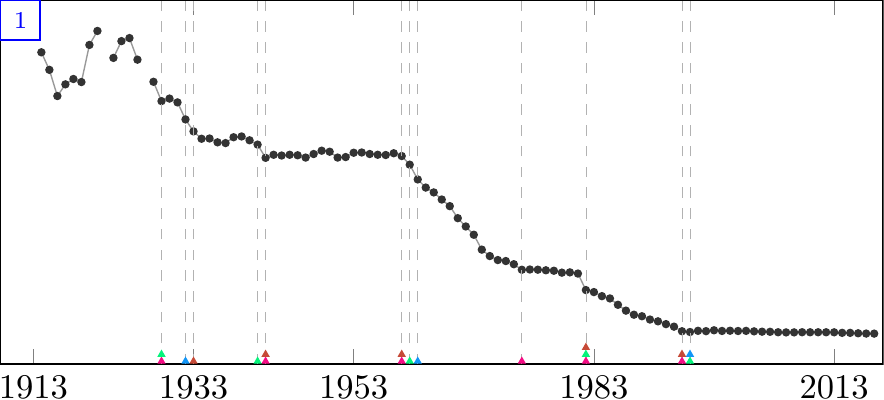}{ukcoalemploy}{uk\_coal\_employ}{Number of workers 
	employed in British coal mines. Data obtained from the 
	U.K.~government.}
\MyFigure{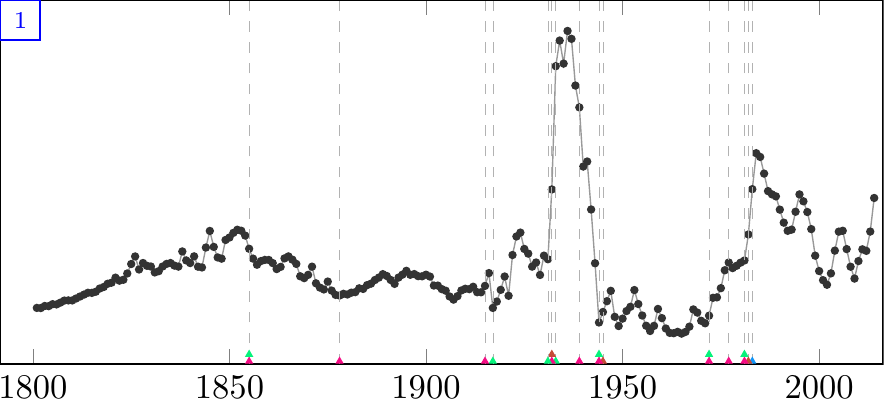}{unemploymentnl}{unemployment\_nl}{Unemployment 
	rates in The Netherlands. Data obtained from Statistics Netherlands.}
\MyFigure{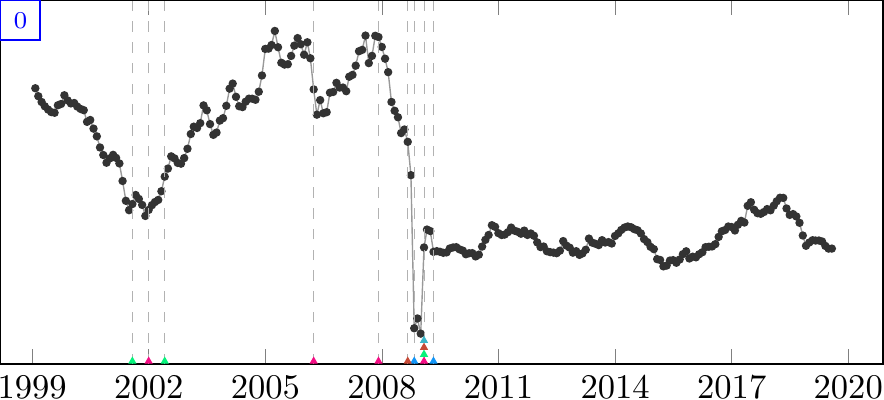}{usdisk}{usd\_isk}{Exchange rate between the US Dollar and 
	the Icelandic Kr\'ona in the years around the financial crisis.  Data 
	obtained from EuroStat.}
\MyFigure{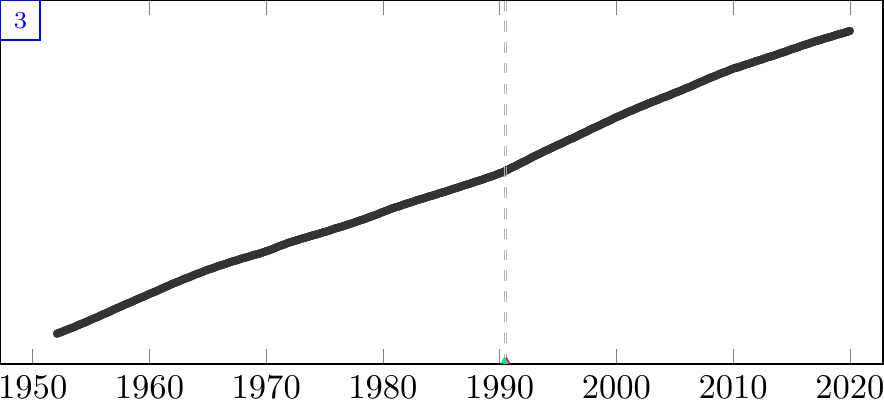}{uspopulation}{us\_population}{Population of the 
	U.S.~over time. Data obtained from the U.S.~Census Bureau.}
\MyFigure{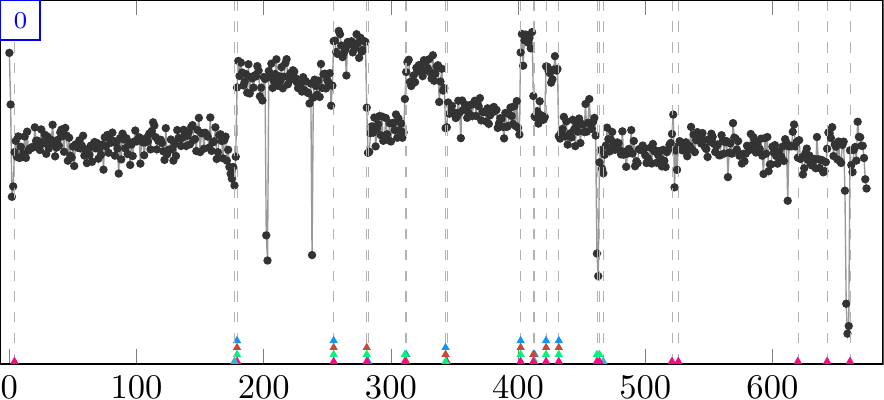}{welllog}{well\_log}{Well-log data from 
	\citet{oruanaidh1996numerical}. The series has been sampled every 6 
	iterations to reduce the length of the series.}

\FloatBarrier

\subsection{Quality Control}%
\label{sub:quality_control}

This section lists the quality control data sets with known change points.

\MyFigure{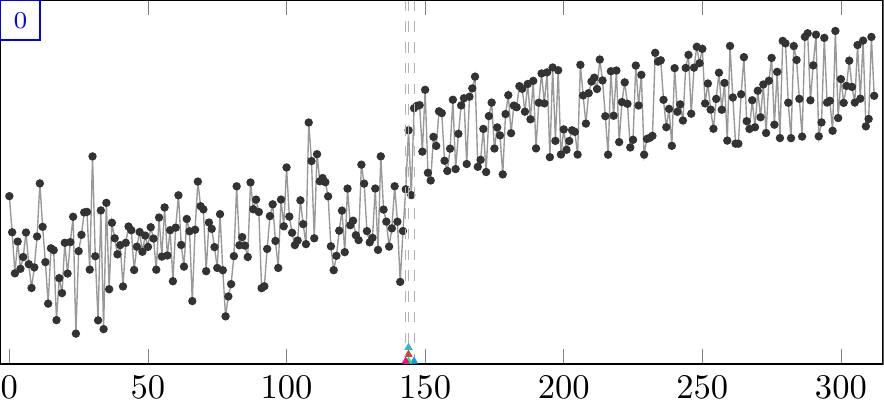}{qualitycontrol1}{quality\_control\_1}{A series 
	with a known change point at index 146. The series has Gaussian noise 
	and a small trend before the change point and an offset and uniform 
	noise after the change point.}
\MyFigure{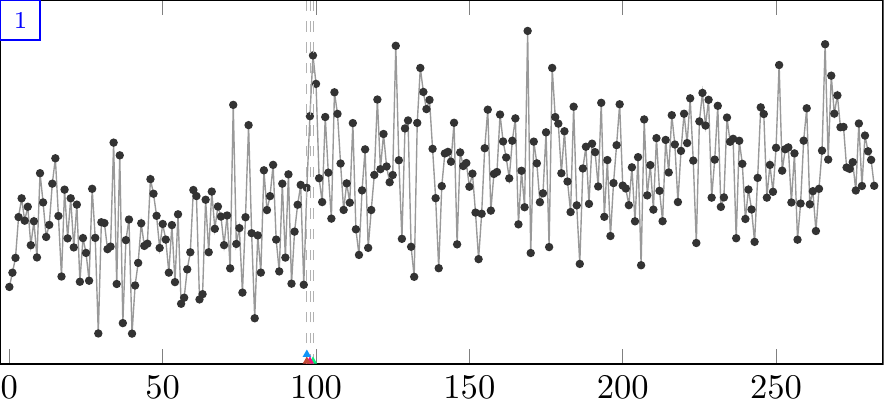}{qualitycontrol2}{quality\_control\_2}{A series 
	with a known change point at index 97. The series has constant noise 
	and a mean shift change point.}
\MyFigure{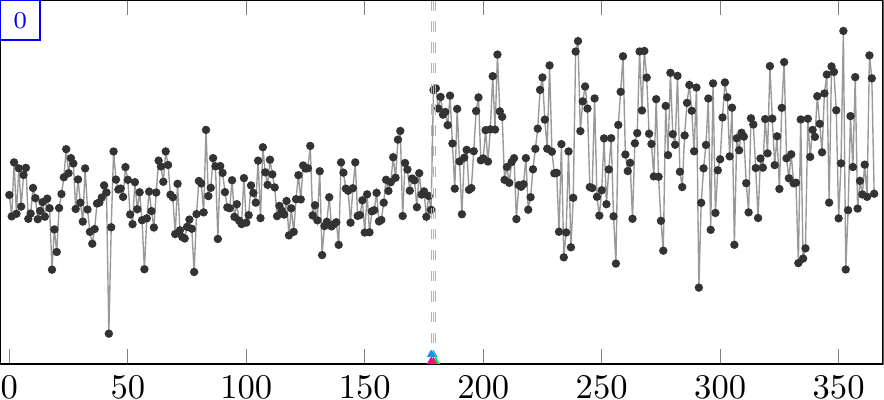}{qualitycontrol3}{quality\_control\_3}{A series 
	with a known change point at index 179. The series has noise 
	$\mathcal{N}(0, 1)$ before the change point and $\mathcal{N}(2, 2)$ 
	after the change point, as well as an outlier at index 42.}
\MyFigure{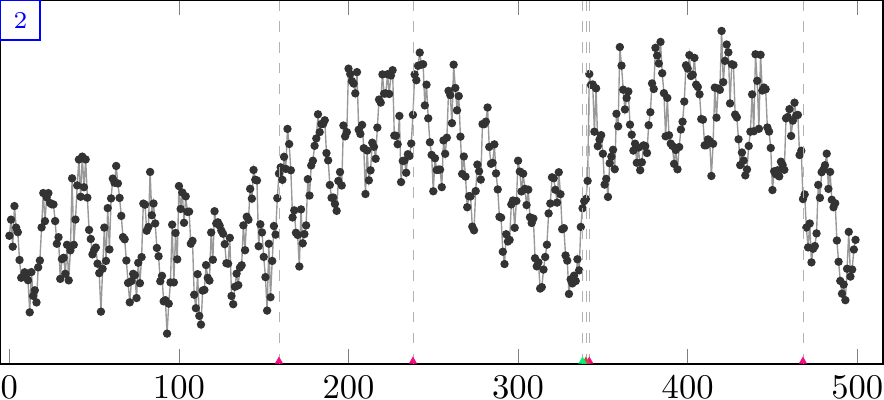}{qualitycontrol4}{quality\_control\_4}{A series 
	with a known change point at index 341. The series has multiple 
	periodic components of different amplitude and a mean shift change 
	point.}
\MyFigure{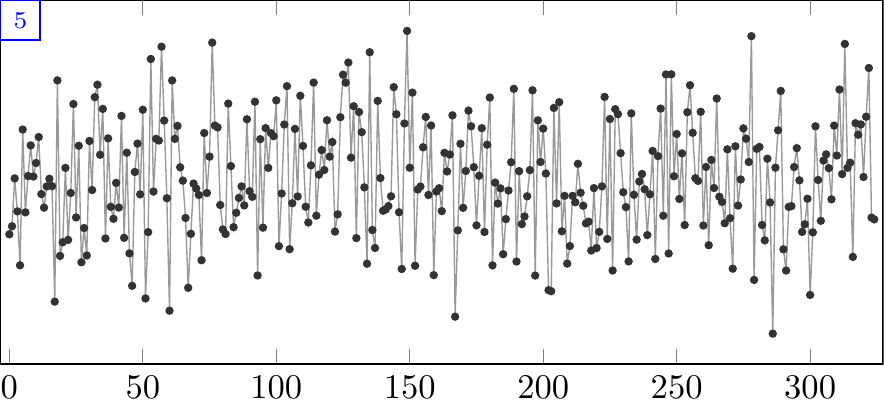}{qualitycontrol5}{quality\_control\_5}{A series of 
	$\mathcal{N}(0, 1)$ noise without a change point.}

\FloatBarrier

\subsection{Introductory Series}%
\label{sub:introduction}

This section lists the synthetic time series shown to the annotators during 
the introduction. Since all annotators had to successfully complete the 
introduction before they could continue with real time series, the annotations 
are not shown for these series.

\MyFigure{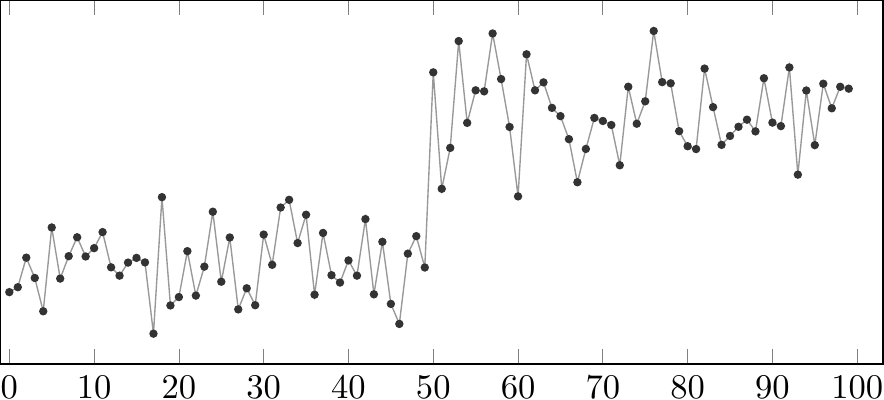}{demo100}{demo\_100}{Series illustrating mean shift change 
	point (index 50).}
\MyFigure{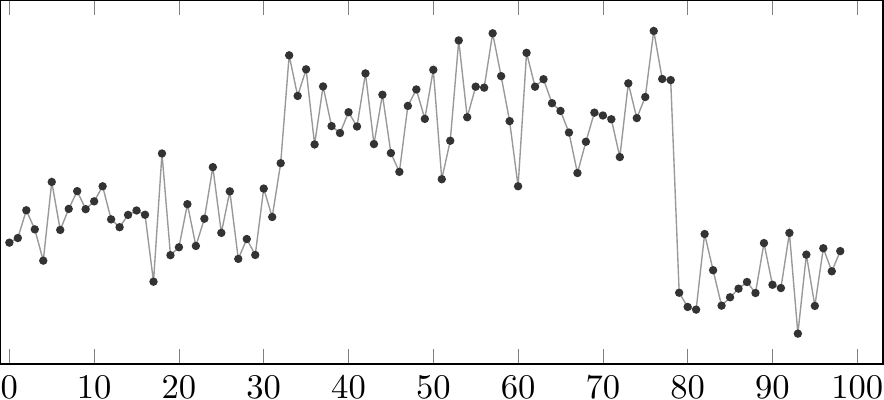}{demo200}{demo\_200}{Series illustrating multiple change 
	points (index 33 and 79).}
\MyFigure{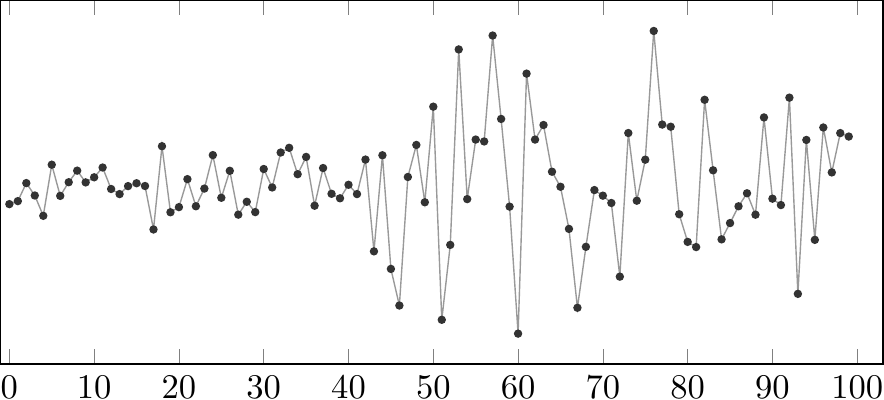}{demo300}{demo\_300}{Series illustrating variance change 
	(index 43).}
\MyFigure{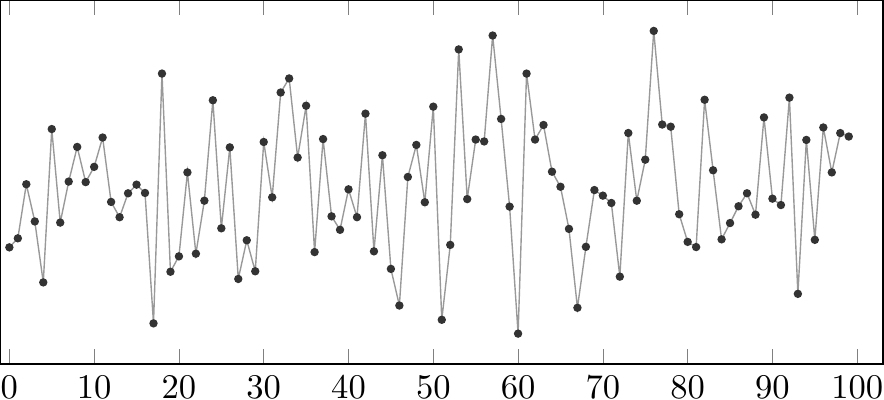}{demo400}{demo\_400}{Series illustrating no change points.}
\MyFigure{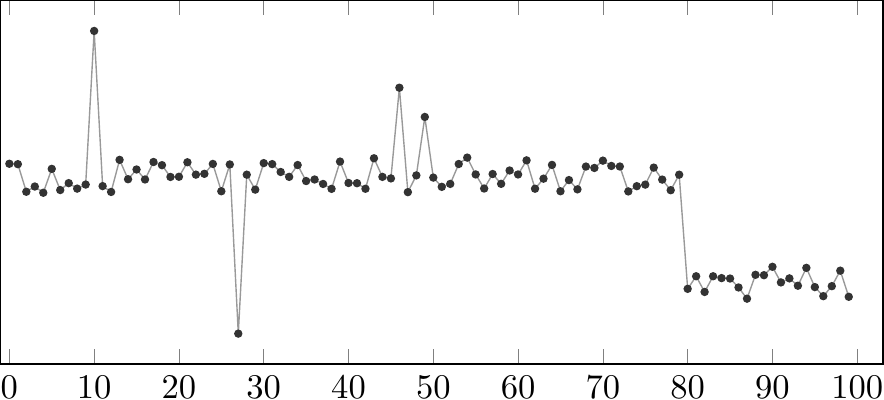}{demo500}{demo\_500}{Series illustrating outliers and a 
	mean shift change point (index 80).}
\MyFigure{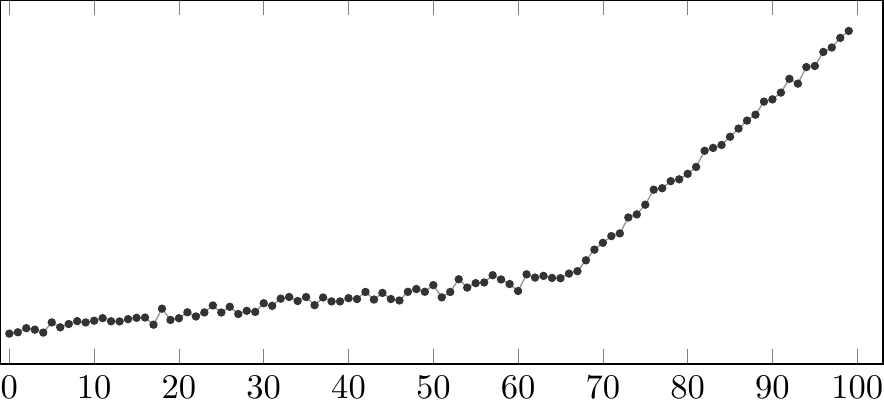}{demo600}{demo\_600}{Series illustrating a trend change 
	(index 65).}
\MyFigure{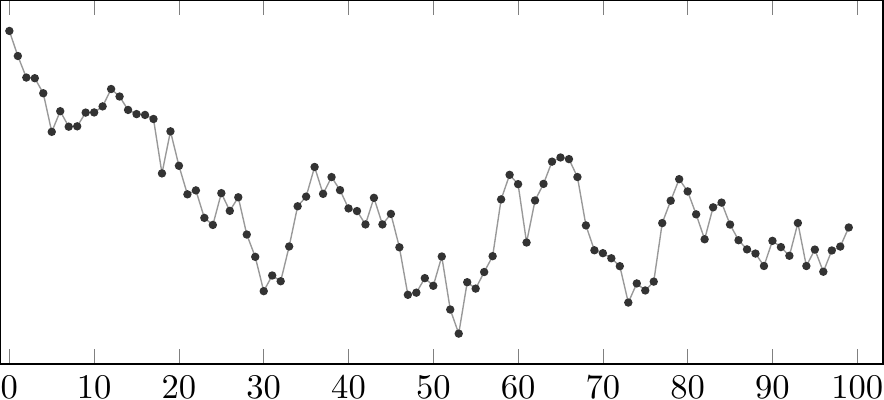}{demo650}{demo\_650}{Series illustrating a random walk (no 
	change point).}
\MyFigure{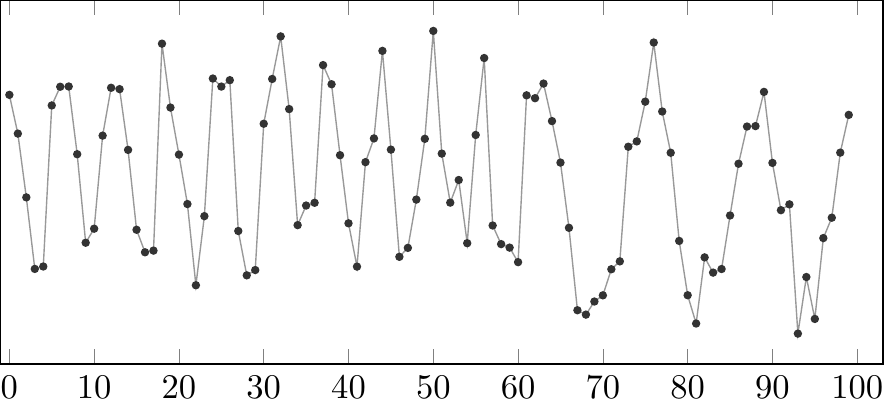}{demo700}{demo\_700}{Series illustrating a change in 
	periodicity (index 57).}
\MyFigure{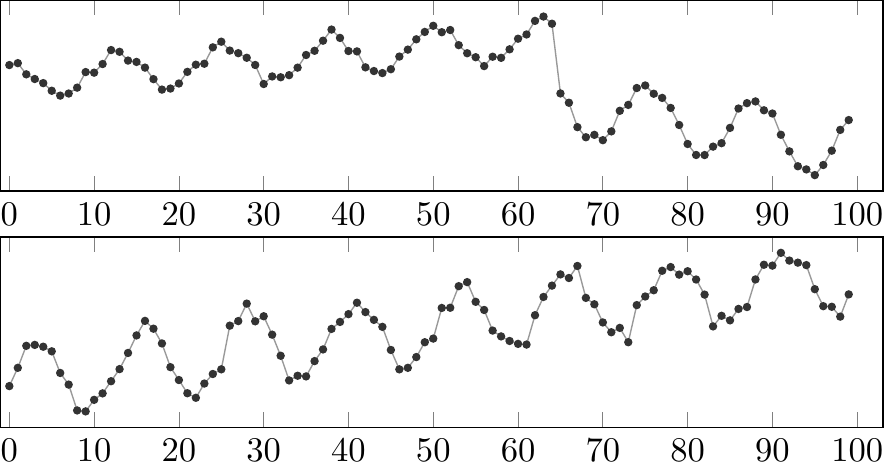}{demo800}{demo\_800}{Series illustrating a multidimensional 
	series with a change point in one dimension (index 65).}

\FloatBarrier
\bibliographystyle{myauthoryear}
\bibliography{references}

\end{document}

%% file: analysis_output/constants/SeriesLengthMean.tex
327.7%

%% file: analysis_output/constants/SeriesLengthMin.tex
15%

%% file: analysis_output/constants/SeriesLengthMax.tex
991%

%% file: analysis_output/constants/UniqueAnnotationsMean.tex
7.4%

%% file: analysis_output/constants/UniqueAnnotationsStd.tex
7.0%

%% file: analysis_output/constants/UniqueAnnotationsMin.tex
0%

%% file: analysis_output/constants/UniqueAnnotationsMax.tex
26%

%% file: analysis_output/tables/aggregate_scores_wide.tex
\begin{tabular}{lrrcrrcrrcrr}
\toprule
 & \multicolumn{5}{c}{Default} &  & \multicolumn{5}{c}{Oracle} \\
\cmidrule(lr){2-6} \cmidrule(lr){8-12}
 & \multicolumn{2}{c}{Univariate} &  & \multicolumn{2}{c}{Multivariate} &  & \multicolumn{2}{c}{Univariate} &  & \multicolumn{2}{c}{Multivariate} \\
\cmidrule(lr){2-3} \cmidrule(lr){5-6} \cmidrule(lr){8-9} \cmidrule(lr){11-12}
 & Cover & F1 & & Cover & F1 & & Cover & F1 & & Cover & F1\\
\midrule
\textsc{amoc}     & \textbf{0.668} & \textbf{0.653} &  &                &                &  & 0.717          & 0.773          &  &                &                \\
\textsc{binseg}   & \textbf{0.672} & \textbf{0.698} &  &                &                &  & \textbf{0.774} & \textbf{0.873} &  &                &                \\
\textsc{bocpd}    & \textbf{0.594} & \textbf{0.662} &  & \textbf{0.455} & \textbf{0.610} &  & \textbf{0.783} & \textbf{0.886} &  & \textbf{0.801} & \textbf{0.941} \\
\textsc{bocpdms}  & \textbf{0.590} & 0.495          &  & \textbf{0.512} & \textbf{0.426} &  & 0.753          & 0.659          &  & \textbf{0.689} & 0.654          \\
\textsc{cpnp}     & 0.488          & \textbf{0.586} &  &                &                &  & \textbf{0.759} & \textbf{0.845} &  &                &                \\
\textsc{ecp}      & 0.470          & 0.560          &  & \textbf{0.402} & \textbf{0.545} &  & 0.693          & 0.773          &  & \textbf{0.590} & \textbf{0.725} \\
\textsc{kcpa}     & 0.069          & 0.124          &  & 0.047          & 0.071          &  & 0.608          & 0.686          &  & \textbf{0.649} & \textbf{0.774} \\
\textsc{pelt}     & \textbf{0.652} & \textbf{0.674} &  &                &                &  & \textbf{0.772} & \textbf{0.864} &  &                &                \\
\textsc{prophet}  & 0.522          & 0.472          &  &                &                &  & 0.554          & 0.502          &  &                &                \\
\textsc{rbocpdms} & 0.561          & 0.397          &  & \textbf{0.485} & \textbf{0.352} &  & 0.717          & 0.677          &  & \textbf{0.649} & 0.559          \\
\textsc{rfpop}    & 0.341          & 0.476          &  &                &                &  & \textbf{0.784} & \textbf{0.870} &  &                &                \\
\textsc{segneigh} & \textbf{0.642} & \textbf{0.635} &  &                &                &  & \textbf{0.777} & \textbf{0.875} &  &                &                \\
\textsc{wbs}      & 0.264          & 0.365          &  &                &                &  & 0.366          & 0.482          &  &                &                \\
\textsc{zero}     & \textbf{0.566} & \textbf{0.645} &  & \textbf{0.464} & \textbf{0.577} &  & 0.566          & 0.645          &  & 0.464          & 0.577          \\
\bottomrule
\end{tabular}

%% file: analysis_output/tables/annotation_simulation_pvalues.tex
\begin{tabular}{lrr}
Dataset & Covering & F1\\
\toprule
\verb+apple+ & 0.014 & 0.001\\
\verb+bank+ & 0.000 & 0.000\\
\verb+bee_waggle_6+ & 0.001 & 0.000\\
\verb+bitcoin+ & 0.008 & 0.049\\
\verb+brent_spot+ & \textbf{0.149} & 0.018\\
\verb+businv+ & 0.011 & 0.000\\
\verb+centralia+ & \textbf{0.172} & \textbf{0.744}\\
\verb+children_per_woman+ & 0.000 & 0.000\\
\verb+co2_canada+ & 0.013 & 0.000\\
\verb+construction+ & 0.016 & 0.008\\
\verb+debt_ireland+ & 0.000 & 0.001\\
\verb+gdp_argentina+ & 0.042 & 0.001\\
\verb+gdp_croatia+ & 0.006 & 0.013\\
\verb+gdp_iran+ & 0.015 & 0.006\\
\verb+gdp_japan+ & 0.009 & 0.000\\
\verb+global_co2+ & \textbf{0.051} & 0.000\\
\verb+homeruns+ & \textbf{0.606} & 0.043\\
\verb+iceland_tourism+ & 0.000 & 0.000\\
\verb+jfk_passengers+ & 0.004 & 0.003\\
\verb+lga_passengers+ & \textbf{0.275} & 0.003\\
\verb+measles+ & 0.000 & 0.000\\
\verb+nile+ & 0.002 & 0.000\\
\verb+occupancy+ & \textbf{0.416} & 0.001\\
\verb+ozone+ & 0.009 & 0.001\\
\verb+quality_control_1+ & 0.000 & 0.000\\
\verb+quality_control_2+ & 0.001 & 0.000\\
\verb+quality_control_3+ & 0.000 & 0.000\\
\verb+quality_control_4+ & \textbf{0.145} & 0.001\\
\verb+quality_control_5+ & 0.000 & 0.000\\
\verb+rail_lines+ & 0.002 & 0.001\\
\verb+ratner_stock+ & 0.000 & 0.012\\
\verb+robocalls+ & 0.019 & 0.018\\
\verb+run_log+ & \textbf{0.052} & 0.001\\
\verb+scanline_126007+ & \textbf{0.735} & 0.014\\
\verb+scanline_42049+ & 0.002 & 0.000\\
\verb+seatbelts+ & 0.005 & 0.000\\
\verb+shanghai_license+ & 0.000 & 0.000\\
\verb+uk_coal_employ+ & \textbf{0.389} & 0.002\\
\verb+unemployment_nl+ & \textbf{0.718} & 0.005\\
\verb+us_population+ & 0.011 & 0.000\\
\verb+usd_isk+ & 0.001 & 0.003\\
\verb+well_log+ & 0.005 & 0.001\\
\bottomrule
\end{tabular}

%% file: analysis_output/tables/descriptive_statistics.tex
\begin{tabular}{llrrrrr}
 & Frequency & Length ($T$) & Dim. ($d$) & Min CP & Max CP & Avg CP\\
\toprule
\verb+ 1. apple+ & Daily$^{\dagger}$ & 622 & 2 & 1 & 8 & 2.4 \\
\verb+ 2. bank+ & Daily & 581 & 1 & 0 & 0 & 0.0 \\
\verb+ 3. bee_waggle_6+ & Unit & 609 & 4 & 0 & 2 & 0.4 \\
\verb+ 4. bitcoin+ & Daily$^{\dagger}$ & 774 & 1 & 1 & 7 & 4.0 \\
\verb+ 5. brent_spot+ & Fortnightly & 500 & 1 & 2 & 11 & 6.0 \\
\verb+ 6. businv+ & Monthly & 330 & 1 & 0 & 3 & 2.2 \\
\verb+ 7. centralia+ & Decenially & 15 & 1 & 0 & 3 & 1.2 \\
\verb+ 8. children_per_woman+ & Yearly & 301 & 1 & 1 & 4 & 2.2 \\
\verb+ 9. co2_canada+ & Yearly & 215 & 1 & 2 & 7 & 4.4 \\
\verb+10. construction+ & Monthly & 319 & 1 & 0 & 2 & 1.2 \\
\verb+11. debt_ireland+ & Yearly & 21 & 1 & 2 & 4 & 2.4 \\
\verb+12. gdp_argentina+ & Yearly & 59 & 1 & 0 & 3 & 1.2 \\
\verb+13. gdp_croatia+ & Yearly & 24 & 1 & 0 & 1 & 0.6 \\
\verb+14. gdp_iran+ & Yearly & 58 & 1 & 0 & 3 & 1.6 \\
\verb+15. gdp_japan+ & Yearly & 58 & 1 & 0 & 1 & 0.4 \\
\verb+16. global_co2+ & Quadrennial & 104 & 1 & 0 & 2 & 0.8 \\
\verb+17. homeruns+ & Yearly & 118 & 1 & 0 & 7 & 2.2 \\
\verb+18. iceland_tourism+ & Monthly & 199 & 1 & 0 & 1 & 0.2 \\
\verb+19. jfk_passengers+ & Monthly & 468 & 1 & 0 & 2 & 1.0 \\
\verb+20. lga_passengers+ & Monthly & 468 & 1 & 0 & 7 & 3.4 \\
\verb+21. measles+ & Weekly & 991 & 1 & 0 & 1 & 0.2 \\
\verb+22. nile+ & Yearly & 100 & 1 & 0 & 1 & 0.6 \\
\verb+23. occupancy+ & Every 16 min. & 509 & 4 & 2 & 12 & 6.2 \\
\verb+24. ozone+ & Yearly & 54 & 1 & 0 & 2 & 1.0 \\
\verb+25. quality_control_1+ & Unit & 313 & 1 & 1 & 1 & 1.0 \\
\verb+26. quality_control_2+ & Unit & 283 & 1 & 0 & 1 & 0.8 \\
\verb+27. quality_control_3+ & Unit & 366 & 1 & 1 & 1 & 1.0 \\
\verb+28. quality_control_4+ & Unit & 500 & 1 & 0 & 4 & 1.2 \\
\verb+29. quality_control_5+ & Unit & 325 & 1 & 0 & 0 & 0.0 \\
\verb+30. rail_lines+ & Yearly & 37 & 1 & 1 & 2 & 1.8 \\
\verb+31. ratner_stock+ & Daily$^{\dagger}$ & 600 & 1 & 1 & 2 & 1.6 \\
\verb+32. robocalls+ & Monthly & 52 & 1 & 0 & 2 & 1.6 \\
\verb+33. run_log+ & Every 5 sec. & 376 & 2 & 0 & 9 & 6.6 \\
\verb+34. scanline_126007+ & Unit & 481 & 1 & 0 & 22 & 5.2 \\
\verb+35. scanline_42049+ & Unit & 481 & 1 & 2 & 10 & 6.6 \\
\verb+36. seatbelts+ & Monthly & 192 & 1 & 0 & 3 & 1.8 \\
\verb+37. shanghai_license+ & Monthly & 205 & 1 & 1 & 2 & 1.2 \\
\verb+38. uk_coal_employ+ & Yearly & 105 & 1 & 0 & 6 & 3.8 \\
\verb+39. unemployment_nl+ & Yearly & 214 & 1 & 0 & 9 & 4.0 \\
\verb+40. us_population+ & Monthly & 816 & 1 & 0 & 1 & 0.4 \\
\verb+41. usd_isk+ & Monthly & 247 & 1 & 1 & 4 & 2.4 \\
\verb+42. well_log+ & Unit & 675 & 1 & 2 & 17 & 9.6 \\
\bottomrule
\end{tabular}

%% file: analysis_output/tables/default_cover_combined_full.tex
\begin{tabular}{lcccccccccccccc}
Dataset & \textsc{amoc} & \textsc{binseg} & \textsc{bocpd} & \textsc{bocpdms} & \textsc{cpnp} & \textsc{ecp} & \textsc{kcpa} & \textsc{pelt} & \textsc{prophet} & \textsc{rbocpdms} & \textsc{rfpop} & \textsc{segneigh} & \textsc{wbs} & \textsc{zero}\\
\toprule
\verb+bank+ & 0.967 & 0.509 & 0.048 & 0.188 & 0.053 & 0.127 & 0.036 & 0.509 & 0.361 & 0.644 & 0.036 & 0.509 & 0.048 & \textbf{1.000}\\
\verb+bitcoin+ & \textbf{0.764} & 0.754 & 0.717 & 0.748 & 0.364 & 0.209 & 0.046 & 0.758 & 0.723 & F & 0.168 & 0.758 & 0.304 & 0.516\\
\verb+brent_spot+ & 0.423 & 0.627 & 0.586 & 0.265 & 0.411 & 0.387 & 0.022 & 0.627 & 0.527 & 0.504 & 0.225 & \textbf{0.630} & 0.242 & 0.266\\
\verb+businv+ & 0.574 & 0.562 & 0.463 & 0.459 & 0.402 & 0.311 & 0.013 & \textbf{0.603} & 0.478 & 0.529 & 0.123 & 0.494 & 0.108 & 0.461\\
\verb+centralia+ & \textbf{0.675} & 0.564 & 0.612 & 0.650 & 0.564 & \textbf{0.675} & 0.440 & 0.564 & \textbf{0.675} & 0.624 & 0.528 & 0.564 & 0.253 & \textbf{0.675}\\
\verb+children_per_woman+ & \textbf{0.798} & \textbf{0.798} & 0.758 & 0.427 & 0.486 & 0.397 & 0.048 & \textbf{0.798} & 0.521 & 0.762 & 0.154 & 0.771 & 0.186 & 0.429\\
\verb+co2_canada+ & 0.527 & 0.608 & \textbf{0.716} & 0.276 & 0.514 & 0.639 & 0.263 & 0.611 & 0.540 & 0.432 & 0.497 & 0.612 & 0.480 & 0.278\\
\verb+construction+ & 0.525 & 0.466 & 0.395 & 0.571 & 0.334 & 0.352 & 0.016 & 0.423 & 0.502 & \textbf{0.584} & 0.092 & 0.423 & 0.198 & 0.575\\
\verb+debt_ireland+ & 0.321 & \textbf{0.635} & 0.584 & 0.312 & \textbf{0.635} & 0.321 & 0.210 & \textbf{0.635} & 0.321 & 0.306 & 0.489 & \textbf{0.635} & 0.248 & 0.321\\
\verb+gdp_argentina+ & 0.667 & 0.667 & 0.630 & 0.711 & 0.451 & \textbf{0.737} & 0.061 & 0.667 & 0.534 & 0.711 & 0.332 & 0.631 & 0.068 & \textbf{0.737}\\
\verb+gdp_croatia+ & 0.605 & 0.605 & 0.642 & 0.655 & 0.642 & \textbf{0.708} & 0.108 & 0.605 & \textbf{0.708} & 0.655 & 0.353 & 0.605 & 0.108 & \textbf{0.708}\\
\verb+gdp_iran+ & 0.484 & 0.477 & 0.520 & 0.580 & 0.448 & \textbf{0.583} & 0.062 & 0.477 & \textbf{0.583} & 0.580 & 0.248 & 0.503 & 0.066 & \textbf{0.583}\\
\verb+gdp_japan+ & 0.654 & 0.654 & 0.578 & 0.645 & 0.522 & \textbf{0.802} & 0.041 & 0.654 & \textbf{0.802} & 0.777 & 0.269 & 0.654 & 0.048 & \textbf{0.802}\\
\verb+global_co2+ & 0.743 & 0.748 & 0.640 & 0.731 & 0.313 & 0.535 & 0.173 & 0.638 & 0.284 & 0.745 & 0.368 & 0.634 & 0.187 & \textbf{0.758}\\
\verb+homeruns+ & \textbf{0.683} & \textbf{0.683} & 0.526 & 0.677 & 0.516 & 0.611 & 0.047 & \textbf{0.683} & 0.574 & 0.559 & 0.354 & 0.590 & 0.312 & 0.511\\
\verb+iceland_tourism+ & 0.764 & 0.764 & 0.595 & 0.616 & 0.383 & 0.382 & 0.017 & 0.764 & 0.453 & 0.869 & 0.293 & 0.650 & 0.512 & \textbf{0.946}\\
\verb+jfk_passengers+ & \textbf{0.839} & \textbf{0.839} & 0.541 & 0.831 & 0.561 & 0.438 & 0.008 & \textbf{0.839} & 0.373 & 0.668 & 0.264 & 0.791 & 0.409 & 0.630\\
\verb+lga_passengers+ & 0.427 & 0.458 & 0.496 & 0.444 & 0.473 & 0.386 & 0.013 & 0.474 & 0.434 & 0.382 & 0.412 & \textbf{0.543} & 0.477 & 0.383\\
\verb+measles+ & \textbf{0.951} & \textbf{0.951} & 0.081 & 0.255 & 0.098 & 0.060 & 0.005 & 0.213 & 0.603 & 0.296 & 0.046 & 0.367 & 0.081 & \textbf{0.951}\\
\verb+nile+ & 0.880 & 0.880 & \textbf{0.888} & 0.870 & 0.880 & 0.857 & 0.032 & 0.880 & 0.758 & 0.753 & 0.880 & 0.880 & 0.880 & 0.758\\
\verb+ozone+ & \textbf{0.701} & 0.600 & 0.602 & 0.560 & 0.441 & 0.574 & 0.070 & 0.635 & 0.574 & 0.614 & 0.309 & 0.627 & 0.070 & 0.574\\
\verb+rail_lines+ & \textbf{0.786} & \textbf{0.786} & 0.767 & 0.416 & 0.738 & 0.428 & 0.103 & \textbf{0.786} & 0.534 & 0.416 & 0.439 & \textbf{0.786} & 0.103 & 0.428\\
\verb+ratner_stock+ & 0.873 & 0.913 & 0.796 & 0.866 & 0.380 & 0.185 & 0.021 & 0.908 & 0.444 & 0.864 & 0.162 & \textbf{0.914} & 0.176 & 0.450\\
\verb+robocalls+ & 0.641 & 0.641 & \textbf{0.808} & 0.622 & 0.677 & 0.601 & 0.069 & 0.641 & 0.601 & 0.607 & 0.447 & 0.760 & 0.069 & 0.601\\
\verb+scanline_126007+ & 0.519 & 0.464 & 0.346 & 0.506 & 0.433 & 0.327 & 0.025 & 0.444 & 0.503 & F & 0.316 & \textbf{0.567} & 0.263 & 0.503\\
\verb+scanline_42049+ & 0.424 & 0.631 & \textbf{0.892} & 0.832 & 0.529 & 0.490 & 0.121 & 0.745 & 0.441 & 0.418 & 0.257 & 0.730 & 0.432 & 0.211\\
\verb+seatbelts+ & 0.683 & \textbf{0.797} & 0.757 & 0.526 & 0.750 & 0.615 & 0.020 & \textbf{0.797} & 0.628 & 0.526 & 0.484 & 0.765 & 0.727 & 0.528\\
\verb+shanghai_license+ & \textbf{0.920} & \textbf{0.920} & 0.856 & 0.616 & 0.474 & 0.518 & 0.020 & \textbf{0.920} & 0.768 & 0.730 & 0.381 & 0.826 & 0.209 & 0.547\\
\verb+uk_coal_employ+ & M & M & M & 0.429 & M & 0.356 & 0.356 & M & \textbf{0.481} & M & M & M & M & 0.356\\
\verb+unemployment_nl+ & 0.507 & \textbf{0.669} & 0.495 & 0.570 & 0.503 & 0.470 & 0.039 & 0.648 & 0.507 & 0.539 & 0.243 & 0.648 & 0.222 & 0.507\\
\verb+us_population+ & 0.736 & 0.391 & 0.219 & 0.801 & 0.391 & 0.089 & 0.003 & 0.506 & 0.096 & 0.632 & \textbf{0.803} & 0.307 & 0.043 & \textbf{0.803}\\
\verb+usd_isk+ & 0.853 & 0.735 & 0.672 & \textbf{0.866} & 0.480 & 0.616 & 0.023 & 0.730 & 0.436 & 0.578 & 0.163 & 0.730 & 0.194 & 0.436\\
\verb+well_log+ & 0.453 & 0.695 & 0.776 & \textbf{0.787} & 0.772 & 0.601 & 0.020 & 0.679 & 0.411 & 0.661 & 0.787 & 0.647 & 0.719 & 0.225\\
\midrule
\verb+quality_control_1+ & \textbf{0.992} & \textbf{0.992} & 0.887 & 0.990 & 0.683 & 0.620 & 0.010 & \textbf{0.992} & 0.693 & 0.986 & 0.655 & \textbf{0.992} & 0.687 & 0.503\\
\verb+quality_control_2+ & 0.922 & 0.922 & \textbf{0.927} & 0.921 & 0.922 & 0.927 & 0.010 & 0.922 & 0.723 & 0.922 & 0.922 & 0.922 & 0.922 & 0.638\\
\verb+quality_control_3+ & 0.996 & 0.996 & \textbf{0.997} & 0.977 & 0.831 & \textbf{0.997} & 0.008 & 0.996 & 0.500 & 0.990 & 0.658 & 0.996 & 0.996 & 0.500\\
\verb+quality_control_4+ & 0.563 & 0.518 & 0.511 & 0.670 & 0.481 & 0.257 & 0.009 & 0.538 & 0.508 & \textbf{0.733} & 0.059 & 0.538 & 0.080 & 0.673\\
\verb+quality_control_5+ & \textbf{1.000} & \textbf{1.000} & \textbf{1.000} & 0.994 & \textbf{1.000} & \textbf{1.000} & 0.006 & \textbf{1.000} & \textbf{1.000} & 0.994 & \textbf{1.000} & \textbf{1.000} & \textbf{1.000} & \textbf{1.000}\\
\midrule
\verb+apple+ &  &  & 0.365 & 0.359 &  & 0.305 & 0.007 &  &  & \textbf{0.694} &  &  &  & 0.425\\
\verb+bee_waggle_6+ &  &  & 0.089 & 0.860 &  & 0.116 & 0.004 &  &  & 0.226 &  &  &  & \textbf{0.891}\\
\verb+occupancy+ &  &  & \textbf{0.549} & 0.443 &  & 0.530 & 0.050 &  &  & 0.437 &  &  &  & 0.236\\
\verb+run_log+ &  &  & \textbf{0.815} & 0.387 &  & 0.657 & 0.128 &  &  & 0.584 &  &  &  & 0.304\\
\bottomrule
\end{tabular}

%% file: analysis_output/tables/default_f1_combined_full.tex
\begin{tabular}{lcccccccccccccc}
Dataset & \textsc{amoc} & \textsc{binseg} & \textsc{bocpd} & \textsc{bocpdms} & \textsc{cpnp} & \textsc{ecp} & \textsc{kcpa} & \textsc{pelt} & \textsc{prophet} & \textsc{rbocpdms} & \textsc{rfpop} & \textsc{segneigh} & \textsc{wbs} & \textsc{zero}\\
\toprule
\verb+bank+ & 0.667 & 0.400 & 0.047 & 0.118 & 0.044 & 0.154 & 0.008 & 0.400 & 0.154 & 0.286 & 0.015 & 0.333 & 0.039 & \textbf{1.000}\\
\verb+bitcoin+ & 0.367 & 0.426 & \textbf{0.692} & 0.269 & 0.463 & 0.338 & 0.092 & 0.672 & 0.446 & F & 0.284 & 0.672 & 0.464 & 0.450\\
\verb+brent_spot+ & 0.272 & 0.483 & 0.521 & 0.239 & \textbf{0.607} & 0.478 & 0.104 & 0.465 & 0.249 & 0.321 & 0.490 & 0.431 & 0.516 & 0.315\\
\verb+businv+ & 0.455 & 0.370 & 0.270 & 0.370 & 0.304 & 0.301 & 0.047 & 0.370 & 0.275 & 0.238 & 0.245 & 0.312 & 0.230 & \textbf{0.588}\\
\verb+centralia+ & 0.763 & 0.909 & 0.909 & 0.846 & 0.909 & 0.763 & 0.714 & 0.909 & 0.763 & 0.846 & \textbf{1.000} & 0.909 & 0.556 & 0.763\\
\verb+children_per_woman+ & 0.618 & 0.618 & \textbf{0.637} & 0.337 & 0.326 & 0.349 & 0.068 & 0.618 & 0.310 & 0.432 & 0.246 & 0.337 & 0.271 & 0.507\\
\verb+co2_canada+ & 0.544 & 0.691 & 0.619 & 0.265 & 0.578 & \textbf{0.817} & 0.169 & 0.661 & 0.482 & 0.381 & 0.569 & 0.661 & 0.520 & 0.361\\
\verb+construction+ & 0.516 & \textbf{0.709} & 0.634 & 0.410 & 0.602 & 0.574 & 0.038 & \textbf{0.709} & 0.324 & 0.291 & 0.185 & \textbf{0.709} & 0.316 & 0.696\\
\verb+debt_ireland+ & 0.469 & 0.760 & 0.760 & 0.611 & 0.760 & 0.469 & 0.519 & 0.760 & 0.469 & 0.530 & \textbf{0.824} & 0.760 & 0.538 & 0.469\\
\verb+gdp_argentina+ & 0.889 & 0.889 & \textbf{0.947} & 0.583 & 0.818 & 0.824 & 0.131 & 0.889 & 0.615 & 0.452 & 0.571 & \textbf{0.947} & 0.148 & 0.824\\
\verb+gdp_croatia+ & 0.583 & 0.583 & \textbf{1.000} & 0.583 & \textbf{1.000} & 0.824 & 0.160 & 0.583 & 0.824 & 0.452 & 0.400 & 0.583 & 0.167 & 0.824\\
\verb+gdp_iran+ & 0.492 & 0.492 & 0.622 & 0.492 & 0.330 & \textbf{0.652} & 0.219 & 0.492 & \textbf{0.652} & 0.395 & 0.609 & 0.395 & 0.246 & \textbf{0.652}\\
\verb+gdp_japan+ & 0.615 & 0.615 & 0.800 & 0.471 & 0.667 & \textbf{0.889} & 0.068 & 0.615 & \textbf{0.889} & 0.471 & 0.190 & 0.615 & 0.077 & \textbf{0.889}\\
\verb+global_co2+ & \textbf{0.929} & \textbf{0.929} & 0.754 & 0.458 & 0.421 & 0.754 & 0.167 & 0.754 & 0.463 & 0.458 & 0.293 & 0.754 & 0.179 & 0.846\\
\verb+homeruns+ & \textbf{0.812} & \textbf{0.812} & 0.729 & 0.552 & 0.604 & 0.675 & 0.133 & \textbf{0.812} & 0.723 & 0.508 & 0.661 & 0.397 & 0.593 & 0.659\\
\verb+iceland_tourism+ & 0.643 & 0.643 & 0.486 & 0.391 & 0.391 & 0.391 & 0.021 & 0.643 & 0.220 & 0.667 & 0.200 & 0.486 & 0.200 & \textbf{0.947}\\
\verb+jfk_passengers+ & \textbf{0.776} & \textbf{0.776} & 0.437 & 0.559 & 0.574 & 0.437 & 0.026 & \textbf{0.776} & 0.354 & 0.296 & 0.344 & 0.650 & 0.437 & 0.723\\
\verb+lga_passengers+ & 0.422 & 0.438 & \textbf{0.630} & 0.297 & 0.606 & 0.392 & 0.054 & 0.438 & 0.366 & 0.348 & 0.498 & 0.395 & 0.524 & 0.535\\
\verb+measles+ & \textbf{0.947} & \textbf{0.947} & 0.069 & 0.124 & 0.118 & 0.069 & 0.004 & 0.124 & 0.391 & 0.117 & 0.030 & 0.327 & 0.039 & \textbf{0.947}\\
\verb+nile+ & \textbf{1.000} & \textbf{1.000} & \textbf{1.000} & 0.800 & \textbf{1.000} & \textbf{1.000} & 0.040 & \textbf{1.000} & 0.824 & 0.452 & \textbf{1.000} & \textbf{1.000} & \textbf{1.000} & 0.824\\
\verb+ozone+ & 0.531 & 0.650 & 0.650 & 0.531 & 0.750 & 0.723 & 0.109 & \textbf{1.000} & 0.723 & 0.559 & 0.375 & \textbf{1.000} & 0.113 & 0.723\\
\verb+rail_lines+ & 0.846 & 0.846 & 0.846 & 0.349 & \textbf{0.966} & 0.537 & 0.200 & 0.846 & 0.423 & 0.349 & 0.571 & 0.846 & 0.205 & 0.537\\
\verb+ratner_stock+ & 0.776 & \textbf{0.824} & 0.783 & 0.500 & 0.306 & 0.282 & 0.034 & 0.650 & 0.280 & 0.559 & 0.203 & \textbf{0.824} & 0.250 & 0.571\\
\verb+robocalls+ & 0.800 & 0.800 & \textbf{0.966} & 0.696 & 0.832 & 0.636 & 0.179 & 0.800 & 0.636 & 0.593 & 0.714 & \textbf{0.966} & 0.182 & 0.636\\
\verb+scanline_126007+ & 0.710 & 0.739 & 0.820 & 0.718 & \textbf{0.889} & 0.732 & 0.101 & 0.759 & 0.644 & F & 0.649 & 0.732 & 0.616 & 0.644\\
\verb+scanline_42049+ & 0.485 & 0.713 & \textbf{0.962} & 0.844 & 0.705 & 0.580 & 0.164 & 0.804 & 0.269 & 0.390 & 0.460 & 0.748 & 0.571 & 0.276\\
\verb+seatbelts+ & 0.474 & \textbf{0.683} & 0.583 & 0.383 & 0.509 & 0.321 & 0.051 & \textbf{0.683} & 0.452 & 0.383 & 0.494 & 0.583 & 0.583 & 0.621\\
\verb+shanghai_license+ & \textbf{0.868} & \textbf{0.868} & 0.713 & 0.491 & 0.437 & 0.698 & 0.048 & \textbf{0.868} & 0.532 & 0.326 & 0.357 & 0.713 & 0.208 & 0.636\\
\verb+uk_coal_employ+ & M & M & M & 0.495 & M & 0.513 & 0.513 & M & \textbf{0.639} & M & M & M & M & 0.513\\
\verb+unemployment_nl+ & 0.566 & \textbf{0.876} & 0.683 & 0.592 & 0.683 & 0.620 & 0.145 & 0.773 & 0.566 & 0.495 & 0.549 & 0.773 & 0.571 & 0.566\\
\verb+us_population+ & \textbf{1.000} & 0.667 & 0.216 & 0.471 & 0.216 & 0.174 & 0.007 & 0.471 & 0.159 & 0.320 & 0.889 & 0.320 & 0.077 & 0.889\\
\verb+usd_isk+ & \textbf{0.785} & 0.657 & 0.609 & 0.678 & 0.504 & 0.661 & 0.079 & 0.657 & 0.489 & 0.282 & 0.390 & 0.657 & 0.513 & 0.489\\
\verb+well_log+ & 0.279 & 0.534 & 0.796 & 0.797 & 0.822 & 0.818 & 0.069 & 0.555 & 0.149 & 0.517 & \textbf{0.923} & 0.485 & 0.724 & 0.237\\
\midrule
\verb+quality_control_1+ & \textbf{1.000} & \textbf{1.000} & 0.800 & 0.667 & 0.667 & 0.667 & 0.025 & \textbf{1.000} & 0.500 & 0.667 & 0.667 & \textbf{1.000} & 0.571 & 0.667\\
\verb+quality_control_2+ & \textbf{1.000} & \textbf{1.000} & \textbf{1.000} & 0.667 & \textbf{1.000} & \textbf{1.000} & 0.028 & \textbf{1.000} & 0.545 & 0.667 & \textbf{1.000} & \textbf{1.000} & \textbf{1.000} & 0.750\\
\verb+quality_control_3+ & \textbf{1.000} & \textbf{1.000} & \textbf{1.000} & 0.667 & 0.571 & \textbf{1.000} & 0.022 & \textbf{1.000} & 0.667 & 0.667 & 0.286 & \textbf{1.000} & \textbf{1.000} & 0.667\\
\verb+quality_control_4+ & 0.810 & \textbf{0.873} & 0.658 & 0.438 & 0.608 & 0.393 & 0.028 & 0.726 & 0.335 & 0.360 & 0.233 & 0.726 & 0.246 & 0.780\\
\verb+quality_control_5+ & \textbf{1.000} & \textbf{1.000} & \textbf{1.000} & 0.500 & \textbf{1.000} & \textbf{1.000} & 0.006 & \textbf{1.000} & \textbf{1.000} & 0.500 & \textbf{1.000} & \textbf{1.000} & \textbf{1.000} & \textbf{1.000}\\
\midrule
\verb+apple+ &  &  & 0.513 & 0.382 &  & 0.513 & 0.029 &  &  & 0.239 &  &  &  & \textbf{0.594}\\
\verb+bee_waggle_6+ &  &  & 0.121 & 0.388 &  & 0.124 & 0.010 &  &  & 0.179 &  &  &  & \textbf{0.929}\\
\verb+occupancy+ &  &  & \textbf{0.807} & 0.461 &  & 0.750 & 0.107 &  &  & 0.329 &  &  &  & 0.341\\
\verb+run_log+ &  &  & \textbf{1.000} & 0.475 &  & 0.792 & 0.139 &  &  & 0.660 &  &  &  & 0.446\\
\bottomrule
\end{tabular}

%% file: analysis_output/tables/oracle_cover_combined_full.tex
\begin{tabular}{lcccccccccccccc}
Dataset & \textsc{amoc} & \textsc{binseg} & \textsc{bocpd} & \textsc{bocpdms} & \textsc{cpnp} & \textsc{ecp} & \textsc{kcpa} & \textsc{pelt} & \textsc{prophet} & \textsc{rbocpdms} & \textsc{rfpop} & \textsc{segneigh} & \textsc{wbs} & \textsc{zero}\\
\toprule
\verb+bank+ & \textbf{1.000} & \textbf{1.000} & \textbf{1.000} & 0.997 & \textbf{1.000} & 0.238 & 0.509 & \textbf{1.000} & \textbf{1.000} & 0.997 & \textbf{1.000} & \textbf{1.000} & 0.053 & \textbf{1.000}\\
\verb+bitcoin+ & 0.771 & 0.771 & \textbf{0.822} & 0.778 & 0.757 & 0.772 & 0.778 & 0.796 & 0.723 & 0.743 & 0.802 & 0.796 & 0.409 & 0.516\\
\verb+brent_spot+ & 0.503 & 0.650 & 0.667 & 0.484 & 0.642 & 0.653 & 0.571 & 0.659 & 0.527 & \textbf{0.709} & 0.669 & 0.659 & 0.310 & 0.266\\
\verb+businv+ & 0.574 & 0.580 & 0.693 & 0.459 & 0.676 & 0.690 & 0.405 & \textbf{0.705} & 0.539 & 0.689 & 0.603 & \textbf{0.705} & 0.154 & 0.461\\
\verb+centralia+ & 0.675 & 0.675 & \textbf{0.753} & 0.650 & 0.675 & \textbf{0.753} & \textbf{0.753} & 0.675 & 0.675 & 0.624 & 0.675 & 0.675 & 0.253 & 0.675\\
\verb+children_per_woman+ & 0.804 & 0.804 & 0.834 & \textbf{0.925} & 0.817 & 0.718 & 0.613 & 0.804 & 0.521 & 0.839 & 0.798 & 0.804 & 0.284 & 0.429\\
\verb+co2_canada+ & 0.527 & \textbf{0.776} & 0.773 & 0.584 & 0.667 & 0.751 & 0.739 & 0.760 & 0.605 & 0.654 & 0.768 & 0.760 & 0.552 & 0.278\\
\verb+construction+ & 0.629 & 0.626 & 0.631 & 0.571 & 0.575 & 0.524 & 0.395 & 0.575 & 0.561 & \textbf{0.744} & 0.732 & 0.626 & 0.300 & 0.575\\
\verb+debt_ireland+ & 0.635 & \textbf{0.859} & 0.814 & 0.803 & 0.777 & 0.798 & 0.747 & 0.779 & 0.321 & 0.689 & \textbf{0.859} & \textbf{0.859} & 0.248 & 0.321\\
\verb+gdp_argentina+ & \textbf{0.737} & \textbf{0.737} & \textbf{0.737} & 0.711 & \textbf{0.737} & \textbf{0.737} & 0.367 & \textbf{0.737} & 0.592 & 0.711 & \textbf{0.737} & \textbf{0.737} & 0.326 & \textbf{0.737}\\
\verb+gdp_croatia+ & 0.708 & 0.708 & 0.708 & 0.675 & 0.708 & 0.708 & 0.623 & 0.708 & 0.708 & 0.708 & \textbf{0.783} & 0.708 & 0.108 & 0.708\\
\verb+gdp_iran+ & 0.583 & 0.583 & 0.583 & 0.719 & 0.583 & 0.583 & 0.505 & 0.583 & 0.583 & \textbf{0.753} & 0.626 & 0.583 & 0.295 & 0.583\\
\verb+gdp_japan+ & \textbf{0.802} & \textbf{0.802} & \textbf{0.802} & 0.777 & \textbf{0.802} & \textbf{0.802} & 0.525 & \textbf{0.802} & \textbf{0.802} & \textbf{0.802} & \textbf{0.802} & \textbf{0.802} & 0.283 & \textbf{0.802}\\
\verb+global_co2+ & \textbf{0.758} & \textbf{0.758} & \textbf{0.758} & 0.745 & \textbf{0.758} & 0.665 & 0.602 & \textbf{0.758} & 0.284 & 0.745 & \textbf{0.758} & \textbf{0.758} & 0.338 & \textbf{0.758}\\
\verb+homeruns+ & 0.694 & \textbf{0.694} & 0.694 & 0.681 & 0.646 & 0.694 & 0.501 & \textbf{0.694} & 0.575 & 0.688 & 0.687 & \textbf{0.694} & 0.407 & 0.511\\
\verb+iceland_tourism+ & \textbf{0.946} & \textbf{0.946} & \textbf{0.946} & 0.936 & \textbf{0.946} & 0.842 & 0.655 & \textbf{0.946} & 0.498 & 0.936 & \textbf{0.946} & \textbf{0.946} & 0.512 & \textbf{0.946}\\
\verb+jfk_passengers+ & 0.844 & 0.845 & 0.842 & 0.855 & 0.845 & 0.807 & 0.563 & 0.839 & 0.373 & 0.627 & \textbf{0.859} & 0.845 & 0.514 & 0.630\\
\verb+lga_passengers+ & 0.434 & 0.541 & 0.563 & 0.512 & 0.534 & \textbf{0.653} & 0.536 & 0.547 & 0.446 & 0.470 & 0.547 & 0.547 & 0.501 & 0.383\\
\verb+measles+ & \textbf{0.951} & \textbf{0.951} & \textbf{0.951} & 0.950 & \textbf{0.951} & 0.105 & 0.400 & \textbf{0.951} & 0.603 & 0.096 & \textbf{0.951} & \textbf{0.951} & 0.084 & \textbf{0.951}\\
\verb+nile+ & \textbf{0.888} & 0.880 & \textbf{0.888} & 0.876 & 0.880 & \textbf{0.888} & \textbf{0.888} & 0.880 & 0.758 & 0.876 & \textbf{0.888} & 0.880 & 0.880 & 0.758\\
\verb+ozone+ & 0.721 & 0.701 & 0.741 & 0.798 & 0.648 & 0.676 & 0.451 & 0.701 & 0.574 & \textbf{0.798} & 0.783 & 0.701 & 0.309 & 0.574\\
\verb+rail_lines+ & 0.786 & 0.786 & 0.789 & 0.767 & \textbf{0.803} & 0.768 & 0.773 & 0.786 & 0.534 & 0.789 & 0.786 & 0.786 & 0.103 & 0.428\\
\verb+ratner_stock+ & 0.874 & \textbf{0.915} & 0.908 & 0.872 & 0.892 & 0.874 & 0.771 & 0.914 & 0.444 & 0.863 & 0.914 & 0.914 & 0.182 & 0.450\\
\verb+robocalls+ & 0.666 & 0.788 & \textbf{0.808} & 0.741 & 0.806 & \textbf{0.808} & \textbf{0.808} & 0.760 & 0.601 & 0.672 & 0.789 & 0.760 & 0.559 & 0.601\\
\verb+scanline_126007+ & 0.634 & 0.633 & 0.631 & \textbf{0.688} & 0.561 & 0.390 & 0.494 & 0.633 & 0.503 & 0.581 & 0.578 & 0.633 & 0.329 & 0.503\\
\verb+scanline_42049+ & 0.425 & 0.868 & \textbf{0.892} & 0.887 & 0.787 & 0.862 & 0.860 & 0.862 & 0.441 & 0.667 & 0.866 & 0.870 & 0.690 & 0.211\\
\verb+seatbelts+ & 0.683 & 0.797 & 0.800 & 0.630 & \textbf{0.825} & 0.800 & 0.800 & 0.797 & 0.635 & 0.658 & 0.797 & 0.797 & 0.727 & 0.528\\
\verb+shanghai_license+ & \textbf{0.930} & \textbf{0.930} & 0.929 & 0.911 & 0.929 & 0.920 & 0.497 & \textbf{0.930} & 0.804 & 0.912 & 0.924 & \textbf{0.930} & 0.351 & 0.547\\
\verb+uk_coal_employ+ & M & M & M & \textbf{0.504} & M & 0.356 & 0.356 & M & 0.481 & M & M & M & M & 0.356\\
\verb+unemployment_nl+ & 0.627 & 0.669 & \textbf{0.673} & 0.656 & 0.652 & 0.501 & 0.476 & 0.650 & 0.507 & 0.634 & 0.650 & 0.650 & 0.447 & 0.507\\
\verb+us_population+ & \textbf{0.803} & \textbf{0.803} & 0.737 & 0.802 & 0.734 & 0.508 & 0.272 & \textbf{0.803} & 0.135 & 0.713 & \textbf{0.803} & \textbf{0.803} & 0.050 & \textbf{0.803}\\
\verb+usd_isk+ & 0.865 & 0.865 & \textbf{0.875} & 0.869 & 0.853 & 0.858 & 0.714 & 0.865 & 0.436 & 0.818 & 0.865 & 0.865 & 0.401 & 0.436\\
\verb+well_log+ & 0.463 & 0.830 & 0.803 & 0.797 & 0.825 & 0.846 & \textbf{0.864} & 0.814 & 0.411 & 0.753 & 0.846 & 0.814 & 0.768 & 0.225\\
\midrule
\verb+quality_control_1+ & \textbf{0.996} & 0.992 & \textbf{0.996} & 0.990 & 0.992 & 0.989 & 0.620 & 0.992 & 0.693 & 0.990 & 0.992 & 0.992 & 0.687 & 0.503\\
\verb+quality_control_2+ & 0.927 & \textbf{0.927} & \textbf{0.927} & 0.922 & 0.922 & 0.927 & \textbf{0.927} & 0.922 & 0.723 & 0.922 & 0.922 & \textbf{0.927} & 0.922 & 0.638\\
\verb+quality_control_3+ & \textbf{0.997} & 0.996 & \textbf{0.997} & 0.991 & 0.996 & \textbf{0.997} & \textbf{0.997} & 0.996 & 0.500 & 0.991 & 0.996 & 0.996 & 0.996 & 0.500\\
\verb+quality_control_4+ & \textbf{0.742} & 0.673 & 0.673 & 0.670 & 0.673 & 0.540 & 0.535 & 0.673 & 0.673 & 0.670 & 0.673 & 0.673 & 0.506 & 0.673\\
\verb+quality_control_5+ & \textbf{1.000} & \textbf{1.000} & \textbf{1.000} & 0.994 & \textbf{1.000} & \textbf{1.000} & \textbf{1.000} & \textbf{1.000} & \textbf{1.000} & 0.994 & \textbf{1.000} & \textbf{1.000} & \textbf{1.000} & \textbf{1.000}\\
\midrule
\verb+apple+ &  &  & \textbf{0.846} & 0.783 &  & 0.758 & 0.462 &  &  & 0.699 &  &  &  & 0.425\\
\verb+bee_waggle_6+ &  &  & \textbf{0.891} & 0.887 &  & 0.116 & 0.730 &  &  & 0.737 &  &  &  & \textbf{0.891}\\
\verb+occupancy+ &  &  & 0.645 & 0.524 &  & \textbf{0.666} & 0.581 &  &  & 0.544 &  &  &  & 0.236\\
\verb+run_log+ &  &  & \textbf{0.824} & 0.563 &  & 0.819 & \textbf{0.824} &  &  & 0.616 &  &  &  & 0.304\\
\bottomrule
\end{tabular}

%% file: analysis_output/tables/oracle_f1_combined_full.tex
\begin{tabular}{lcccccccccccccc}
Dataset & \textsc{amoc} & \textsc{binseg} & \textsc{bocpd} & \textsc{bocpdms} & \textsc{cpnp} & \textsc{ecp} & \textsc{kcpa} & \textsc{pelt} & \textsc{prophet} & \textsc{rbocpdms} & \textsc{rfpop} & \textsc{segneigh} & \textsc{wbs} & \textsc{zero}\\
\toprule
\verb+bank+ & \textbf{1.000} & \textbf{1.000} & \textbf{1.000} & 0.500 & \textbf{1.000} & 0.200 & 0.333 & \textbf{1.000} & \textbf{1.000} & 0.500 & \textbf{1.000} & \textbf{1.000} & 0.043 & \textbf{1.000}\\
\verb+bitcoin+ & 0.507 & 0.780 & 0.733 & 0.533 & 0.611 & 0.625 & 0.665 & \textbf{0.783} & 0.446 & 0.422 & 0.727 & \textbf{0.783} & 0.690 & 0.450\\
\verb+brent_spot+ & 0.465 & 0.670 & 0.631 & 0.309 & \textbf{0.735} & 0.636 & 0.553 & 0.641 & 0.249 & 0.621 & 0.641 & 0.651 & 0.564 & 0.315\\
\verb+businv+ & 0.588 & 0.588 & \textbf{0.650} & 0.455 & 0.588 & 0.370 & 0.294 & 0.588 & 0.275 & 0.553 & 0.588 & 0.588 & 0.289 & 0.588\\
\verb+centralia+ & 0.909 & \textbf{1.000} & \textbf{1.000} & \textbf{1.000} & \textbf{1.000} & 0.909 & \textbf{1.000} & \textbf{1.000} & 0.763 & 0.974 & \textbf{1.000} & \textbf{1.000} & 0.556 & 0.763\\
\verb+children_per_woman+ & 0.678 & 0.678 & 0.787 & \textbf{0.826} & 0.734 & 0.551 & 0.525 & 0.778 & 0.310 & 0.580 & 0.637 & 0.778 & 0.500 & 0.507\\
\verb+co2_canada+ & 0.544 & 0.893 & \textbf{0.924} & 0.479 & 0.677 & 0.875 & 0.867 & 0.894 & 0.482 & 0.621 & 0.894 & 0.894 & 0.681 & 0.361\\
\verb+construction+ & 0.696 & \textbf{0.709} & \textbf{0.709} & 0.516 & \textbf{0.709} & \textbf{0.709} & 0.634 & \textbf{0.709} & 0.324 & 0.615 & \textbf{0.709} & \textbf{0.709} & 0.523 & 0.696\\
\verb+debt_ireland+ & 0.760 & \textbf{1.000} & \textbf{1.000} & 0.892 & 0.958 & 0.980 & \textbf{1.000} & \textbf{1.000} & 0.469 & 0.958 & \textbf{1.000} & \textbf{1.000} & 0.538 & 0.469\\
\verb+gdp_argentina+ & 0.889 & \textbf{0.947} & \textbf{0.947} & 0.583 & 0.889 & 0.889 & 0.800 & \textbf{0.947} & 0.615 & 0.818 & \textbf{0.947} & \textbf{0.947} & 0.421 & 0.824\\
\verb+gdp_croatia+ & \textbf{1.000} & 0.824 & \textbf{1.000} & 0.583 & \textbf{1.000} & 0.824 & 0.800 & 0.824 & 0.824 & 0.824 & \textbf{1.000} & 0.824 & 0.167 & 0.824\\
\verb+gdp_iran+ & 0.696 & 0.696 & 0.862 & 0.838 & 0.667 & 0.824 & 0.734 & 0.808 & 0.652 & \textbf{0.899} & 0.833 & 0.808 & 0.576 & 0.652\\
\verb+gdp_japan+ & \textbf{1.000} & \textbf{1.000} & \textbf{1.000} & 0.615 & \textbf{1.000} & \textbf{1.000} & 0.500 & \textbf{1.000} & 0.889 & \textbf{1.000} & \textbf{1.000} & \textbf{1.000} & 0.222 & 0.889\\
\verb+global_co2+ & 0.929 & 0.929 & 0.929 & 0.458 & 0.846 & \textbf{0.929} & 0.750 & 0.929 & 0.463 & 0.634 & 0.929 & 0.929 & 0.250 & 0.846\\
\verb+homeruns+ & 0.812 & 0.889 & 0.829 & 0.675 & 0.681 & 0.829 & 0.829 & 0.873 & 0.723 & 0.879 & \textbf{0.961} & 0.873 & 0.664 & 0.659\\
\verb+iceland_tourism+ & 0.947 & 0.947 & 0.947 & 0.486 & 0.947 & \textbf{1.000} & 0.486 & 0.947 & 0.220 & 0.667 & 0.947 & 0.947 & 0.200 & 0.947\\
\verb+jfk_passengers+ & \textbf{0.776} & \textbf{0.776} & \textbf{0.776} & 0.650 & \textbf{0.776} & 0.651 & 0.651 & \textbf{0.776} & 0.354 & 0.420 & \textbf{0.776} & \textbf{0.776} & 0.437 & 0.723\\
\verb+lga_passengers+ & 0.561 & 0.620 & 0.715 & 0.563 & 0.634 & \textbf{0.892} & 0.598 & 0.602 & 0.366 & 0.561 & 0.666 & 0.620 & 0.674 & 0.535\\
\verb+measles+ & \textbf{0.947} & \textbf{0.947} & \textbf{0.947} & 0.486 & \textbf{0.947} & 0.080 & 0.281 & \textbf{0.947} & 0.391 & 0.090 & \textbf{0.947} & \textbf{0.947} & 0.041 & \textbf{0.947}\\
\verb+nile+ & \textbf{1.000} & \textbf{1.000} & \textbf{1.000} & 0.800 & \textbf{1.000} & \textbf{1.000} & \textbf{1.000} & \textbf{1.000} & 0.824 & 0.800 & \textbf{1.000} & \textbf{1.000} & \textbf{1.000} & 0.824\\
\verb+ozone+ & 0.776 & \textbf{1.000} & 0.966 & 0.778 & \textbf{1.000} & \textbf{1.000} & 0.667 & \textbf{1.000} & 0.723 & \textbf{1.000} & \textbf{1.000} & \textbf{1.000} & 0.286 & 0.723\\
\verb+rail_lines+ & 0.846 & \textbf{0.966} & \textbf{0.966} & 0.889 & \textbf{0.966} & \textbf{0.966} & 0.800 & 0.846 & 0.537 & 0.730 & 0.889 & \textbf{0.966} & 0.205 & 0.537\\
\verb+ratner_stock+ & 0.776 & 0.832 & \textbf{0.929} & 0.559 & 0.776 & 0.776 & 0.754 & 0.824 & 0.280 & 0.686 & 0.824 & 0.824 & 0.378 & 0.571\\
\verb+robocalls+ & 0.800 & \textbf{0.966} & \textbf{0.966} & 0.750 & \textbf{0.966} & \textbf{0.966} & \textbf{0.966} & \textbf{0.966} & 0.636 & 0.765 & \textbf{0.966} & \textbf{0.966} & 0.714 & 0.636\\
\verb+scanline_126007+ & 0.710 & 0.920 & \textbf{0.964} & 0.837 & 0.927 & 0.870 & 0.838 & 0.889 & 0.644 & 0.570 & 0.914 & 0.940 & 0.818 & 0.644\\
\verb+scanline_42049+ & 0.485 & 0.931 & \textbf{0.962} & 0.921 & 0.809 & 0.910 & 0.908 & 0.910 & 0.269 & 0.526 & 0.910 & 0.920 & 0.650 & 0.276\\
\verb+seatbelts+ & 0.824 & \textbf{0.838} & 0.683 & 0.583 & \textbf{0.838} & 0.683 & 0.683 & 0.683 & 0.452 & 0.621 & 0.683 & \textbf{0.838} & 0.583 & 0.621\\
\verb+shanghai_license+ & \textbf{0.966} & \textbf{0.966} & 0.868 & 0.651 & \textbf{0.966} & 0.868 & 0.465 & \textbf{0.966} & 0.532 & 0.778 & \textbf{0.966} & \textbf{0.966} & 0.385 & 0.636\\
\verb+uk_coal_employ+ & M & M & M & 0.617 & M & 0.513 & 0.513 & M & \textbf{0.639} & M & M & M & M & 0.513\\
\verb+unemployment_nl+ & 0.742 & \textbf{0.889} & 0.876 & 0.797 & 0.773 & 0.755 & 0.744 & 0.816 & 0.566 & 0.697 & 0.851 & 0.816 & 0.801 & 0.566\\
\verb+us_population+ & \textbf{1.000} & \textbf{1.000} & \textbf{1.000} & 0.615 & 0.947 & 0.471 & 0.276 & \textbf{1.000} & 0.159 & 0.320 & 0.889 & \textbf{1.000} & 0.113 & 0.889\\
\verb+usd_isk+ & 0.785 & 0.785 & 0.876 & 0.678 & 0.785 & 0.785 & 0.601 & 0.785 & 0.489 & \textbf{0.881} & 0.785 & 0.785 & 0.636 & 0.489\\
\verb+well_log+ & 0.336 & 0.944 & 0.908 & 0.796 & 0.878 & 0.928 & 0.950 & 0.912 & 0.149 & 0.644 & \textbf{0.966} & 0.912 & 0.832 & 0.237\\
\midrule
\verb+quality_control_1+ & \textbf{1.000} & \textbf{1.000} & \textbf{1.000} & 0.667 & \textbf{1.000} & \textbf{1.000} & 0.667 & \textbf{1.000} & 0.500 & 0.667 & \textbf{1.000} & \textbf{1.000} & 0.667 & 0.667\\
\verb+quality_control_2+ & \textbf{1.000} & \textbf{1.000} & \textbf{1.000} & 0.800 & \textbf{1.000} & \textbf{1.000} & \textbf{1.000} & \textbf{1.000} & 0.750 & 0.667 & \textbf{1.000} & \textbf{1.000} & \textbf{1.000} & 0.750\\
\verb+quality_control_3+ & \textbf{1.000} & \textbf{1.000} & \textbf{1.000} & 0.766 & \textbf{1.000} & \textbf{1.000} & \textbf{1.000} & \textbf{1.000} & 0.667 & 0.667 & \textbf{1.000} & \textbf{1.000} & \textbf{1.000} & 0.667\\
\verb+quality_control_4+ & 0.810 & \textbf{0.873} & 0.810 & 0.561 & 0.810 & 0.726 & 0.658 & 0.810 & 0.780 & 0.438 & \textbf{0.873} & 0.810 & 0.608 & 0.780\\
\verb+quality_control_5+ & \textbf{1.000} & \textbf{1.000} & \textbf{1.000} & 0.500 & \textbf{1.000} & \textbf{1.000} & \textbf{1.000} & \textbf{1.000} & \textbf{1.000} & 0.500 & \textbf{1.000} & \textbf{1.000} & \textbf{1.000} & \textbf{1.000}\\
\midrule
\verb+apple+ &  &  & \textbf{0.916} & 0.552 &  & 0.745 & 0.634 &  &  & 0.575 &  &  &  & 0.594\\
\verb+bee_waggle_6+ &  &  & \textbf{0.929} & 0.651 &  & 0.233 & 0.651 &  &  & 0.359 &  &  &  & \textbf{0.929}\\
\verb+occupancy+ &  &  & 0.919 & 0.714 &  & \textbf{0.932} & 0.812 &  &  & 0.605 &  &  &  & 0.341\\
\verb+run_log+ &  &  & \textbf{1.000} & 0.698 &  & 0.990 & \textbf{1.000} &  &  & 0.698 &  &  &  & 0.446\\
\bottomrule
\end{tabular}